\documentclass[10pt,twocolumn,letterpaper]{article}

\usepackage[pagenumbers]{iccv} 

\definecolor{iccvblue}{rgb}{0.21,0.49,0.74}
\usepackage[pagebackref,breaklinks,colorlinks,allcolors=iccvblue]{hyperref}

\usepackage{lipsum}
\usepackage{ulem}
\usepackage{graphicx}
\usepackage{array}
\usepackage{xcolor}
\usepackage{multirow} 
\usepackage{algorithm}
\usepackage{algorithmic}
\usepackage{subcaption}
\usepackage{amsthm}

\DeclareMathOperator*{\argmax}{arg\,max}

\usepackage{hhline}
\usepackage{booktabs}
\usepackage{tabularx}
\usepackage{wrapfig}
\usepackage{makecell}
\usepackage{colortbl}

\usepackage[accsupp]{axessibility}

\def\confName{ICCV}
\def\confYear{2025}

\title{Understanding Flatness in Generative Models: Its Role and Benefits}

\author{
\textbf{Taehwan Lee}$^{1,}$\thanks{Equal contribution} \and
\textbf{Kyeongkook Seo}$^{1,\ast}$ \and  
\textbf{Jaejun Yoo}$^{1,2,}$\thanks{Co-corresponding author} \and 
\textbf{Sung Whan Yoon}$^{1,2,\dagger}$ \\
$^{1}$Graduate School of Artificial Intelligence, Ulsan National Institute of Science and Technology
\\
$^{2}$Department of Electrical Engineering, Ulsan National Institute of Science and Technology \\
{\tt\small \{taehwan,kyeongkookseo,jaejun.yoo,shyoon8\}@unist.ac.kr}
}

\newcommand{\ours}[1]{XXX}

\newcommand{\Aref}[1]{Appendix \ref{#1}}

\usepackage{amssymb}
\newtheorem{definition}{Definition}
\newtheorem{proposition}{Proposition}
\newtheorem{theorem}{Theorem}

\newtheorem{corollary}{Corollary}
\newtheorem{remark}{Remark}
\usepackage{comment}

\begin{document}
\maketitle
\begin{abstract}
Flat minima, known to enhance generalization and robustness in supervised learning, remain largely unexplored in generative models.
In this work, we systematically investigate the role of loss surface flatness in generative models, both theoretically and empirically, with a particular focus on diffusion models.
We establish a theoretical claim that flatter minima improve robustness against perturbations in target prior distributions, leading to benefits such as reduced exposure bias---where errors in noise estimation accumulate over iterations---and significantly improved resilience to model quantization, preserving generative performance even under strong quantization constraints. 
We further observe that Sharpness-Aware Minimization (SAM), which explicitly controls the degree of flatness, effectively enhances flatness in diffusion models even surpassing the indirectly promoting flatness methods---Input Perturbation (IP) which enforces the Lipschitz condition, ensembling-based approach like Stochastic Weight Averaging (SWA) and Exponential Moving Average (EMA)---are less effective.
Through extensive experiments on CIFAR-10, LSUN Tower, and FFHQ, we demonstrate that flat minima in diffusion models indeed improve not only generative performance but also robustness.

\end{abstract}    
\section{Introduction}
\label{sec:intro}

\textit{What does it mean for a generative model to have a flat loss landscape?}
While flat minima have been extensively studied in supervised learning, where they are known to enhance generalization and robustness to distribution shifts, e.g., promoting stable label predictions under input shifts such as domain changes~\cite{SWA'18,SWAD'21,SAM'21,ASAM'21,LPF-SGD'22,R-SAM'22,F-SAM'24}, their role in generative models remains largely unexplored. 
\begin{figure}[t!]
    \centering
    \includegraphics[width=0.45\linewidth]{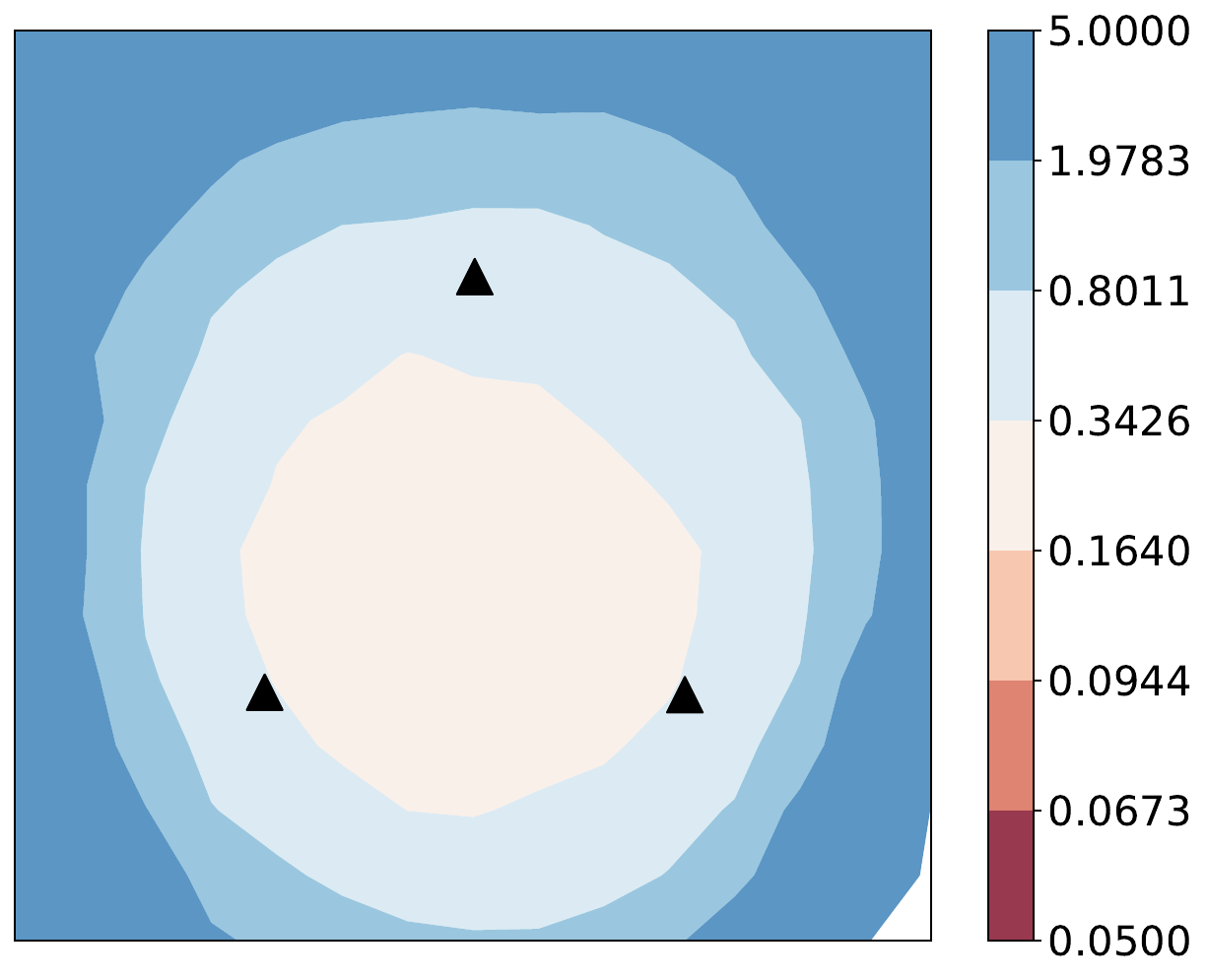}
    \hfill
    \includegraphics[width=0.45\linewidth]{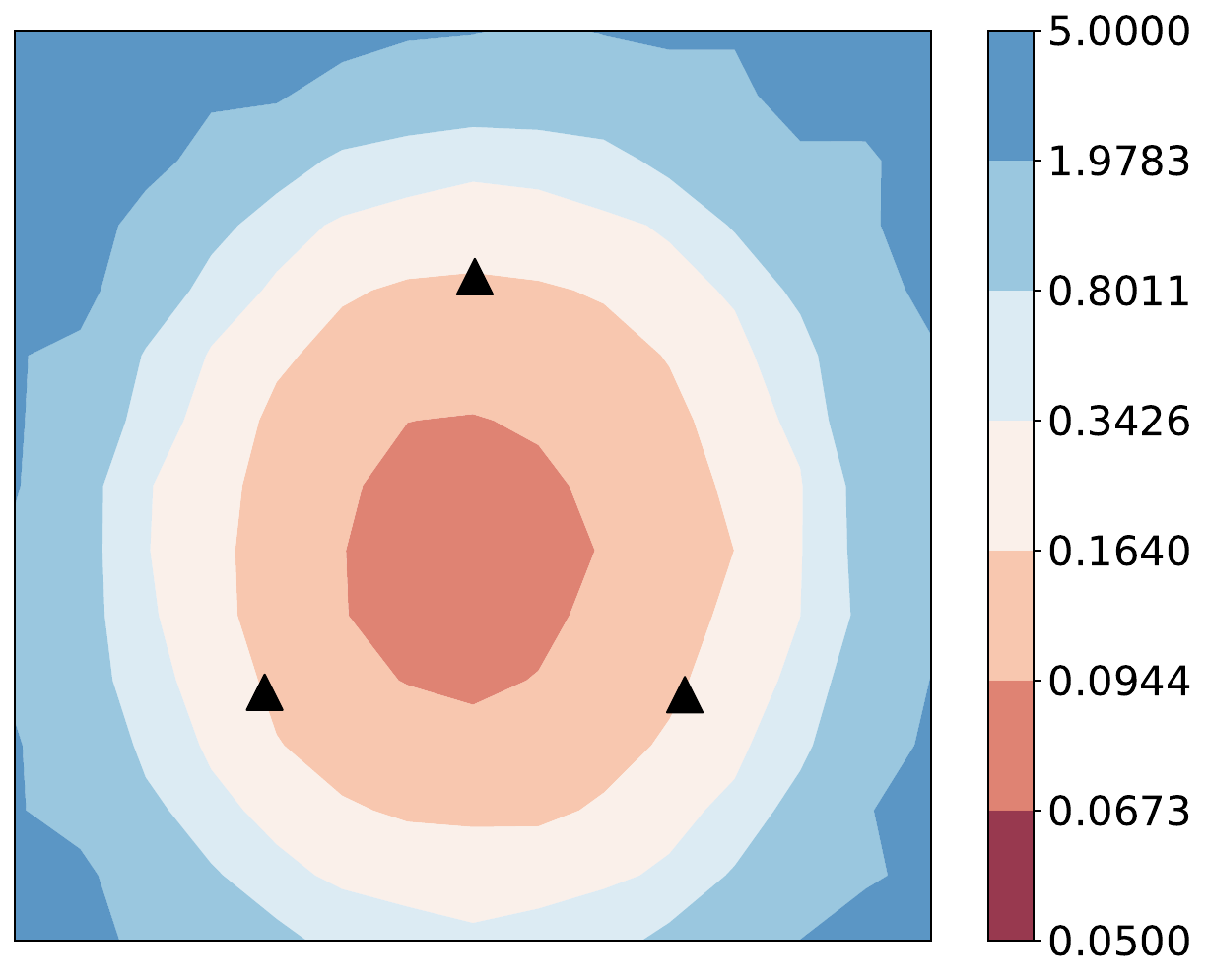}
    \vspace{-2.5mm}
    \caption{Loss surface of ADM (\textit{left}) and ADM+\texttt{SAM} (\textit{right})}
    \label{Fig: Loss landscape for ADM+EMA and +SAM+EMA}
\end{figure}
\begin{table}[t!]
\vspace{-3mm}
    \centering
    \begin{tabular}{c |  c c c}
    \toprule
         &  FID $\downarrow$ (32 $\rightarrow$ 8)& $\|\epsilon_{\theta}\|^2
         $ gap $\downarrow$ & LPF $\downarrow$ \\ \hline
         ADM  & 34.47 $\overset{\text{{\tiny\textcolor{red}{+13.65}}}}{\longrightarrow}$ 48.02 &  +11.39 & 0.097 \\
         \texttt{+SAM} & \textbf{9.01} $\overset{\text{{\tiny\textcolor{green!60!black}{-0.07}}}}{\longrightarrow}$ \textbf{8.94} & \textbf{+3.32}  & \textbf{0.063} \\
    \bottomrule
    \end{tabular}
    \vspace{-2.5mm}
    \caption{Comparison of the 1) Fr\'{e}chet Inception Distance (FID) scores under 32- and 8-bit precision with (\textcolor{red}{+}/\textcolor{green!60!black}{-}) changes, 2) the gap of $\|\epsilon\|_2$ with training, 3) flatness metric called LPF for ADM and \texttt{+SAM} in the CIFAR-10 experiments. $\downarrow$: a lower value is preferred.}
    \label{Tab: Performance for ADM+EMA and +SAM+EMA}
\end{table}
\begin{figure}[t!]
\vspace{-3.5mm}
    \centering
    \begin{subfigure}{0.48\columnwidth}
        \centering
        \includegraphics[width=\columnwidth]{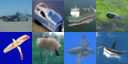}
        \caption{ADM}
        \label{fig:first}
    \end{subfigure}
    \hfill
    \begin{subfigure}{0.48\columnwidth}
        \centering
        \includegraphics[width=\columnwidth]{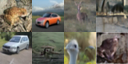}
        \caption{ADM\texttt{+SAM}}
        \label{fig:first}
    \end{subfigure}
    \vspace{-3mm}
    \caption{Generated samples of 8-bit quantized models for CIFAR-10. Model parameters are directly converted from 32-bit to 8-bit after training, without further optimization or adaptation. }
    \label{fig:appetizer}
    \vspace{-5mm}
\end{figure}

At first glance, one might expect that flatness would be beneficial in generative models as well. However, generative models---particularly those based on diffusion processes---operate under fundamentally different principles: rather than mapping structured inputs to labels, they take noise as input and iteratively refine it into coherent outputs through a learned denoising process. This key difference motivates a critical question: \textit{Do flat minima compel the generative model to produce similar contents regardless of input variations?} If so, such behavior would contradict the goal of generative diversity and potentially impair generalization. If not, it remains unclear how flatness affects generative modeling, and if it is indeed beneficial, how can we induce it?

As a preview of our analysis, \Cref{Fig: Loss landscape for ADM+EMA and +SAM+EMA}, \ref{fig:appetizer} and \Cref{Tab: Performance for ADM+EMA and +SAM+EMA}
hint at the role and benefits of the generative models with flat minima. 
We find that SAM~\cite{SAM'21}, an optimizer designed to promote flatter loss surfaces, significantly enhances flatness when applied to a baseline, i.e., Ablated Diffusion Model (ADM)~\cite{ADM}, as shown in \cref{Fig: Loss landscape for ADM+EMA and +SAM+EMA,Tab: Performance for ADM+EMA and +SAM+EMA} (measured using Low-Pass-Filter (LPF) flatness metric \cite{LPF-SGD'22}).
Regarding the benefits from flatness, we observe improved FID, reduced exposure biases (as indicated by the $\|\epsilon\|^2$ gap~\cite{DDPM-ES}), and lower degradation from model quantization (as reflected in 8-bit precision FID and qualitative results in \cref{fig:appetizer}).

Returning to the key question, we set out to investigate the implication of flat loss landscapes in generative models, particularly diffusion models~\cite{DDPM}. To make progress, we first seek to refine our core research question: \textit{``Why might flatter minima be desirable in generative models, and what theoretical guarantees can we establish?"} 

To address these questions, we conduct a theoretical and empirical analysis of flat minima in diffusion models. First, we establish a theoretical link between the model’s parameter space and data probability density (\textbf{\cref{thm1}}), showing that flatter minima improve robustness to perturbations in the target prior distribution. As a consequence, we prove that the divergence between the true distribution and the learned distribution is upper-bounded, indicating that flat minima enable diffusion models to generalize not only to the training distribution but also to variations in perturbed data densities (\textbf{\cref{thm2}}). 
This insight suggests that flat diffusion models exhibit superior robustness when out-of-distribution is given, demonstrating their ability to cope with the error accumulation problem during the iterative sampling process, known as \textit{exposure bias problem}~\cite{DDPM-IP, DDPM-ES, DDPM-TS}.
Moreover, the flat loss landscape in the parameter space directly implies the robustness against model parameter changes, which indicates that the flatness offers the robustness of generative performance when the model has been quantized.

To complement our theoretical findings, we empirically examine the effects of explicit flatness regularization. Initially, we expect that applying SAM to diffusion models would yield improvements similar to those observed in discriminative settings. However, an intriguing discovery emerges: diffusion models are inherently much flatter than anticipated. Unlike discriminative models, where typical SAM trade-off values significantly impact on flatness, applying standard regularization strengths to diffusion models produced almost no discernible change. Further analysis reveals that diffusion models naturally exhibit a flat loss landscape, likely due to the diverse range of noise levels they are trained to denoise. Building on this observation, we find that significantly stronger regularization is required in diffusion models for SAM to induce meaningful flatness, setting them apart from their discriminative counterparts.

To validate these findings, we conduct extensive experiments on CIFAR-10, LSUN Tower, and FFHQ. Our results show that flat minima in diffusion models consistently improve generative performance with enhanced generalization. This improved generalization offers several practical advantages. For instance, we observe that flat diffusion models better mitigate exposure bias, where errors in noise estimation accumulate over iterations, leading to more stable sampling dynamics. We also show that flat diffusion models result in a low quantization error where the model precision is reduced from 32-bit to 8-bit.

We summarize our contributions as follows. Our work provides the first systematic investigation into the role of flat minima in generative models, offering both theoretical insights and practical implications:
\begin{itemize}
\item We explore diffusion models through the lens of flat loss landscapes, revealing their inherent flatness.
\item We establish a theoretical link between flat minima and generalization, proving that flatness improves robustness via an upper-bound on the divergence between the target and learned distributions.
\item We empirically analyze IP~\cite{DDPM-IP}, SWA~\cite{SWA'18}, and EMA, showing that while they improve FID, they do not significantly enhance flatness. In contrast, SAM reshapes the loss surface and effectively promotes flatness in diffusion models.
\item We demonstrate how these insights address practical challenges in generative modeling through extensive evaluations on CIFAR-10, LSUN Tower, and FFHQ, analyzing FID performance, exposure bias, model quantization error, and loss landscape properties.
\end{itemize}

\section{Preliminaries and Backgrounds}
\label{sec:related}

\subsection{Optimization for flatness} 

\begin{algorithm}[t!]
	\caption{Pseudo algorithm for SWA, EMA, SAM}
	\label{Flat-minima-searching-Algoritms}
	\textbf{Input}: $w_0$: Initial weights, \textcolor{red}{$\rho$}: size of perturbation,\\
    \textcolor{blue}{$c$}: Cycle length,
    \textcolor{YellowOrange}{$\lambda$}: update momentum\\
	\textbf{Output}: $w, w_{SWA}, w_{EMA}$
	\begin{algorithmic}[1] 
		\STATE $w, w_{SWA}, w_{EMA} \leftarrow w_0$\{Initialize weights with $w_0$\}
		\FOR{$i = 0, ..., n-1$}
			\IF {\textcolor{blue}{SWA}}
				\STATE $\alpha \leftarrow \alpha(i)$ \{Calculate LR for the iteration\}		
			\ENDIF
			\IF {\textcolor{red}{SAM}}
				\STATE $\hat{w} \leftarrow w+ \rho \frac{\nabla \mathcal{L}_i (w)}{\| \nabla \mathcal{L}_i (w)\|}$ \{Ascent step\}
				\STATE $ w \leftarrow w - \alpha  \nabla \mathcal{L}_i (\hat {w})$ \{Descent step\}
			\ELSE
				\STATE $w \leftarrow w-\alpha \nabla \mathcal{L}_i (w)$
			\ENDIF
			
			\IF {\textcolor{blue}{SWA} \&  mod($i,c$)=0}
				\STATE $n_{models} \leftarrow i/c$ \{Number of models\}
				\STATE $w_{SWA} \leftarrow \frac{w_{SWA}\cdot n_{models}+w}{n_{models}+1}$ \{Update average\}
			\ELSIF {\textcolor{YellowOrange}{EMA}}
				\STATE $w_{EMA} \leftarrow (1- \lambda)\cdot w_{EMA} + \lambda \cdot w  $ \{Update average\}
			\ENDIF

		\ENDFOR
	\end{algorithmic}
\end{algorithm}

\noindent
\textbf{EMA.} EMA is a weight-averaging method that blends a fraction of newly updated parameters with a fraction of previously accumulated parameters at each step. Formally, for parameters $\theta$, the EMA update is:
$\theta_{\text{EMA}} \leftarrow (1 - \lambda) \theta_{\text{EMA}} + \lambda\,\theta, \quad \lambda \in (0,1).$
Heuristically ensembling different models from multiple iterations reaches flat local minima and improves generalization capacities \cite{IterAvg'20, EMA, SwitchEMA}.

\smallskip
\noindent
\textbf{SWA.} SWA~\cite{SWA'18} stabilizes training by maintaining an ongoing average of model weights across training epochs. Rather than relying on the final parameters from a single run, SWA updates this average—typically during the latter part of training—leading to an overall smoothing effect in the loss landscape. Although SWA and EMA share the idea of parameter averaging, SWA is directly motivated by the search for flatter minima, which improves generalization.

\smallskip
\noindent
\textbf{SAM}. SAM~\cite{SAM'21} is an optimization technique that promotes generalization by steering model parameters toward flatter regions of the loss landscape. It incorporates a sharpness term into the objective, defined as the maximal increase in loss within an $\ell_2$-ball of radius $\rho$ around the current parameters $\theta$:
$\max_{\|p\|_2 \leq \rho} \big[L(\theta + p) - L(\theta)\big].$
By minimizing this sharpness term alongside the original loss, SAM encourages solutions that are robust to small perturbations in the parameters, thus improving generalization.
The pseudocode for SWA, EMA, and SAM is provided in \Cref{Flat-minima-searching-Algoritms}.

\noindent
\textbf{IP. IP~\cite{DDPM-IP}} was introduced to address the exposure bias problem, which inherently arises in the auto-regressive denoising process of diffusion models. They define exposure bias as a generalization failure to unseen inputs, caused by the accumulation of prediction errors over timesteps. To mitigate this issue, authors propose a smoothed diffusion model that satisfies a Lipschitz condition by training on randomly perturbed input data. This strategy aligns with the flat minima perspective, implicitly promoting generalization to nearby unseen inputs.

\subsection{Score-based Generative Models}\label{SGM-Pre}
Score-based generative models (SGMs)~\cite{SGM1, SGM2, SGM3, SGM_SDE} leverage the score function of a data distribution to iteratively generate samples by solving Stochastic Differential Equations (SDEs). 
Let $p_{\text{data}}(\mathbf{x})$ be the unknown data distribution. The \textit{score function} is defined as the gradient of the log-density:
\begin{equation}
    \nabla_{\mathbf{x}} \log p_{\text{data}}(\mathbf{x}).
\end{equation}
Instead of modeling $p_{\text{data}}(\mathbf{x})$ explicitly, SGMs learn a neural network $s_{\theta}(\mathbf{x}, t)$ to approximate the score function under slight noise perturbations, $t$ is timestep.

SGMs can be formulated as a stochastic process where noise is gradually added to the data in the forward direction and removed in the backward direction.

\smallskip
\noindent
\textbf{Forward SDE.} The forward diffusion process is defined by the following SDE:
\begin{equation} \label{Eq: Forward SDE}
    d\mathbf{x} = f(\mathbf{x}, t) dt + g(t) d\mathbf{w},
\end{equation}
where $f(\mathbf{x}, t)$ is the drift coefficient that controls deterministic changes, $g(t)$ is the diffusion coefficient that controls noise intensity, and $d\mathbf{w}$ denotes a standard Wiener process.

\smallskip
\noindent
\textbf{Reverse SDE.} Using the learned score function, we can reverse the diffusion process to generate new samples. The backward SDE is given by:
\begin{equation} \label{Eq: Backward SDE}
    d\mathbf{x} = \left[ f(\mathbf{x}, t) - g^2(t) \nabla_{\mathbf{x}} \log p_t(\mathbf{x}) \right] dt + g(t) d\bar{\mathbf{w}},
\end{equation}
where $\nabla_{\mathbf{x}} \log p_t(\mathbf{x})$ is the time-dependent score function, approximated by $s_{\theta}(\mathbf{x}, t)$, and $d\bar{\mathbf{w}}$ is a standard Wiener process running in reverse time.
To successfully estimate the score function, the score matching objective is defined as:
\begin{equation} \label{Eq: Score matching objective}
    \mathcal{L}_{\text{SGM}}=\mathbb{E}_{t}\left[\lambda(t) \cdot \mathbb{E}_{p_t(\mathbf{x})}
    \left[ \left\| s_{\theta}(\mathbf{x}, t)  - \nabla_{\mathbf{x}_t} \log p_t(\mathbf{x})\right\|_2^2 \right] \right].
\end{equation}

\noindent
\textbf{Diffusion models.} Diffusion models (DPMs)~\cite{DDPM_nonequilibrium_thermodynamics, DDPM} can be viewed as a discrete-time approximation of SGMs, with the following forward and backward diffusion process:
\begin{align} \label{Eq: diffusion forward distribution}
    q(\mathbf{x}_{t}|\mathbf{x}_{t-1}) &:= \mathcal{N}(\mathbf{x}_t;\sqrt{1-\beta_{t}}\mathbf{x}_{t-1}, \beta_t \mathbf{I}), \\
    q(\mathbf{x}_{t-1}|\mathbf{x}_{t}) &= \mathcal{N}(\mathbf{x}_{t-1};\tilde{\mu}(\mathbf{x}_t, \mathbf{x}_0), \tilde{\beta}_t \mathbf{I}),\label{Eq: diffusion forward posterior} \\
    \tilde{\mu}(\mathbf{x}_t, \mathbf{x}_0) &= \frac{\sqrt{\bar{\alpha}_{t-1}} \beta_t}{1 - \bar{\alpha}_t} \mathbf{x}_0 + \frac{\sqrt{\alpha_t} (1 - \bar{\alpha}_{t-1})}{1 - \bar{\alpha}_t} \mathbf{x}_t,\label{Eq: diffusion tilde_mu tilde_beta}\\
    \tilde{\beta}_t &= \frac{1 - \bar{\alpha}_{t-1}}{1 - \bar{\alpha}_t} \beta_t
\end{align}
where $\beta_t$ is a predefined parameter by variance scheduling, $\alpha_t:=1-\beta_t$, and $\bar{\alpha}_t:=\Pi_{s=1}^t\alpha_s$.

\noindent
\textbf{Exposure bias.} Previous works~\cite{DDPM-IP, DDPM-ES, DDPM-TS} have raised an input discrepancy problem between the training and sampling phases, known as \textit{exposure bias problem}. Specifically, during training, the DPM is provided with ground truth noisy images, whereas during sampling, it receives a denoised image that came from the previous timestep. The presence of errors in these inputs, amplified through the iterative denoising process, makes it challenging for the diffusion model to accurately predict noise. As these errors accumulate over iterative timesteps, they cause the generated samples to significantly deviate from the desired trajectory.

\section{Theoretical Analysis on Flatness}
\label{sec:Theoretical}

We here theoretically analyze the effect of flat minima in SGMs, with an emphasis on diffusion models.
\subsection{Notations}
Our analysis is built upon the settings of SGM, which is a fundamental form of diffusion models. Beyond the brief preliminaries in \Cref{SGM-Pre}, we introduce additional notations and formulations to be used in our analysis. 

Following prior theoretical analysis~\cite{GeneralizationDiffModel}, we formulate the simplified score model $s_{\theta}(\cdot, \cdot)$ as a random feature model consisting of an encoder and a decoder.
\begin{equation} \label{Eq: score model}
    s_{\boldsymbol{\theta}}(\mathbf{x}, t):=\frac{1}{m}\mathbf{\boldsymbol{\theta}}\sigma\big(\mathbf{W}^\top\mathbf{x}+\mathbf{U}^\top\mathbf{e}_{t}\big),
\end{equation}
where $\mathbf{x}\in\mathbb{R}^{d\times 1}$ is the given input, $\mathbf{e}_{t}$ is the timestep embedding for $t$, a group of parameters includes $\boldsymbol{\theta}\in\mathbb{R}^{d\times m}$, $\mathbf{W}\in\mathbb{R}^{d\times m}$ and $\mathbf{U}\in\mathbb{R}^{d_{e}\times m}$, and a few positive integers include $d$, $m$, and $d_e$.
By following the training procedures of diffusion models, $\boldsymbol{\theta}$ is the learnable parameter, while others, i.e., $\mathbf{W}$ and $\mathbf{U}$, which respectively embed $\mathbf{x}$ and $\mathbf{e}_t$, are set to be frozen.

For a given timestep $t$, the score matching loss objective defined in \cref{Eq: Score matching objective}
\begin{equation} \label{eq:sgm}
    \mathcal{L}(\mathbf{x};\boldsymbol{\theta},t,p_t) := \|s_{\boldsymbol{\theta}}(\mathbf{x},t)-\nabla_{\mathbf{x}}\log{p_t(\mathbf{x})}\|_{2}^{2},
\end{equation}
where 
$p_t(\mathbf{x})$ is the probability density function for the target distribution at time $t$.

\subsection{Mathematical Claims}
For the main claims, we follow the formulation of Equation \eqref{eq:sgm}, while omitting timestep $t$ without loss of generality. Our mathematical claims are valid for all timesteps.
We start from defining flat minima in SGMs (\textbf{Definition \ref{def:flat}}), and the robustness against the gap of ground truth prior $p(\mathbf{x})$ and the estimated prior $\hat{p}(\mathbf{x})$ (\textbf{Definition \ref{def:expbias}}). 
For the flat minima definition, we rehearse the definition of flat minima for the supervised learning case~\cite{OvercomingForget'21} and extend it to the $\Delta$-flat minima of SGMs.
\begin{definition}\label{def:flat}
(\textit{$\Delta$-flat minima}) Let us consider a SGM with loss function $\mathcal{L}(\mathbf{x};\boldsymbol{\theta},p)$. A minimum $\boldsymbol{\theta^{*}}$ is $\Delta$-flat minima when the following constraints are hold:
\begin{gather}
\forall \: \delta\in\mathbb{R}^{d\times m} \text{ s.t. } \|\delta\|_{2}\leq\Delta, \:\: \mathcal{L}(\mathbf{x};\boldsymbol{\theta}^{*}+\delta,p) = l^* \nonumber\\
\exists \: \delta\in\mathbb{R}^{d\times m}
\text{ s.t. } \|\delta\|_{2}>\Delta, \:\:
\mathcal{L}(\mathbf{x};\boldsymbol{\theta}^*+\delta,p) > l^*, \nonumber
\end{gather}
where $l^*:=\mathcal{L}(\mathbf{x};\boldsymbol{\theta}^*,p)$ and $\Delta\in\mathbb{R}^{+}$.\footnote{$\forall$ means ``for all," $\exists$ means ``there exists," and $\mathbb{R}^+$ indicates the set of positive real numbers.}
\end{definition}

To extend to robustness against the erroneous estimation of prior probability,
we further define the $\mathcal{E}$-distribution gap robustness of SGMs, where $\mathcal{E}$ indicates the divergence between the ground truth $p$ and the estimated $\hat{p}$:
\begin{definition}\label{def:expbias}
(\textit{$\mathcal{E}$-distribution gap robustness}) 
A minimum $\boldsymbol{\theta^{*}}$ is $\mathcal{E}$-distribution gap robust when the following constraints are hold:
\begin{gather}
\forall \: \hat{p}(\mathbf{x}) \text{ s.t. } D(p||\hat{p}) \leq \mathcal{E}, \:\: \mathcal{L}(\mathbf{x};\boldsymbol{\theta}^{*},\hat{p}) = l^* \nonumber\\
\exists \: \hat{p}(\mathbf{x}) \text{ s.t. } D(p||\hat{p}) > \mathcal{E}, \:\: \mathcal{L}(\mathbf{x};\boldsymbol{\theta}^{*},\hat{p}) > l^*, \nonumber
\end{gather}
where $D(\cdot||\cdot)$ is the divergence between two probability density functions, $\hat{p}$ is the perturbed prior distribution of $\mathbf{x}$, and $\mathcal{E}$ is a positive real number.
\end{definition}

Let us then bridge the flatness and the robustness.
First, based on the definitions of flat minima (\textbf{Definition \ref{def:flat}}) and distributional gap robustness (\textbf{Definition \ref{def:expbias}}), we try to formulate how the perturbation of the parameter, i.e., $\boldsymbol{\theta}+\delta$, links to the perturbed prior distribution $\hat{p}$, with the given equality as follows:
\begin{equation}\label{loss_equality}
\mathcal{L}(\mathbf{x};\boldsymbol{\theta}+\delta,p) = \mathcal{L}(\mathbf{x};\boldsymbol{\theta},\hat{p}).
\end{equation}
This means how the parameter perturbation is translated into the shift of the prior, while keeping the loss unchanged.
Let us provide a mathematical claim for it:

\begin{theorem}\label{thm1}
(A perturbed distribution)
For a given prior distribution of $p(\mathbf{x})$ and the $\delta$-perturbed minimum, i.e., $\boldsymbol{\theta}+\delta$, the following $\hat{p}(\mathbf{x})$ satisfies the equality \eqref{loss_equality}:
\begin{gather}
\hat{p}(\mathbf{x}) = e^{-I(\mathbf{x},\delta)} p(\mathbf{x}),
\end{gather}
where $I(\mathbf{x},\delta):= \frac{1}{2}\mathbf{x}^\top(\delta
\mathbf{W}^\top)\mathbf{x} + \mathbf{x}^\top\delta
(\mathbf{U}^\top\mathbf{e})+C$ with $\delta \mathbf{W}^\top$ being symmetric, and $C\in\mathbb{R}$ is set to satisfy  $\displaystyle\int_{\mathbb{R}^{d}} e^{-I(\mathbf{x},\delta)}p(\mathbf{x})d\mathbf{x}=1$.
\end{theorem}
For analytical convenience, 
 \textbf{\cref{thm1}} derives a perturbed density under symmetric $\delta \mathbf W^\top$, showing that parameter perturbation translates into a ``scaling'' of probability density. Note that $C$ absorbs the normalizing constant.
However, the scaling depends on various parameters. Thus, it is non-trivial to imagine the behavior of $\hat{p}$.
Let us narrow our view to the diffusion model, where it predicts a noise signal sampled from a Gaussian distribution, i.e., $\mathcal{N}(0, \mathbf{I})$. It changes our focus on handling the noise distribution, i.e., $\epsilon$, rather than the distribution of data side, i.e., $p$.
The following corollary describes how the perturbed $\hat{\epsilon}$ looks like:
\begin{corollary}\label{cor1.1}
(Diffusion version of \textbf{\cref{thm1}})
For a given prior Gaussian distribution of noise $\epsilon\sim\mathcal{N}(0,\mathbf{I})$ and the $\delta$-perturbed minimum, i.e., $\boldsymbol{\theta}+\delta$, the following $\hat{\epsilon}$ satisfies the equality \eqref{loss_equality}:
\begin{gather}
\hat{\epsilon} = e^{-I(\mathbf{x},\delta)} \epsilon = \mathcal{N}(\boldsymbol{\mu}_{\delta},\Sigma_\delta),
\end{gather}
where $\Sigma_{\delta}:= 
\bigg(\mathbf {I} + \displaystyle\frac{\delta \mathbf{W}^\top}{m}\bigg)^{-1}$, $\boldsymbol \mu_{\delta}:=\displaystyle\frac{1}{m}\Sigma_{\delta}\delta\mathbf{U}^\top \mathbf{e}$.
\end{corollary}
\begin{remark}
\textbf{($\hat{\epsilon}$ becomes perturbed Gaussian)} For the diffusion models, we emphasize that the perturbation in the parameter space, $\delta$, leads to the perturbation of distribution, which follows the Gaussian distribution.
\end{remark}

Next, let us consider all possible perturbations within the ball of norm, i.e., $\|\delta\|_2\leq \Delta$ and $\delta \mathbf{W}^\top= \mathbf{W}\delta^\top$. The resulting set of perturbed distributions induced by the parameter perturbation is defined as follows:

\begin{definition}\label{def:P}
(A set of perturbed distributions)
For a given distribution of $p(\mathbf{x})$, a set of distributions $\hat{\mathcal{P}}(\mathbf{x};p,\Delta)$ is defined as the set of perturbed distributions $\hat{p}(\mathbf{x})$:
\begin{align}
\hat{\mathcal{P}}&(\mathbf{x};p,\Delta) \nonumber \\ &:= \left\{ e^{-I(\mathbf{x},\delta)} p(\mathbf{x})  \middle|  \|\delta\|_{2} \leq \Delta,  \delta \mathbf{W}^\top = \mathbf{W} \delta^\top\right\}.
\end{align}
\end{definition}

From the definition, $\Delta$-flat minimum flattens all loss values of the perturbed parameters. Consequently, the flat minimum achieves flat loss values for all possible distributions within $\hat{\mathcal{P}}$, as formally shown in the following proposition:

\begin{proposition}\label{prop1}
(A link from $\Delta$-flatness to $\hat{\mathcal{P}}$) A $\Delta$-flat minimum $\boldsymbol{\theta}^*$ achieves the flat loss values for all distributions sampled from the set of perturbed distributions:
\begin{gather}
\forall \: p\sim\hat{\mathcal{P}}(\mathbf{x};p,\Delta), \:\: \mathcal{L}(\mathbf{x};\boldsymbol{\theta}^*,p) = l^* \\
\exists \: p\nsim\hat{\mathcal{P}}(\mathbf{x};p,\Delta), \:\: \mathcal{L}(\mathbf{x};\boldsymbol{\theta}^*,p) > l^*.
\end{gather}
\end{proposition}
We finally link flatness to the distributional gap, i.e., connecting \textbf{\cref{def:flat}} and \textbf{\ref{def:expbias}}. Let us anticipate the distribution in $\hat{\mathcal{P}}$ with the maximal divergence from $p$. The distribution gap $\mathcal{E}$ is then upper bounded by the divergence of the most outreached distribution:

\begin{figure}[t!]
     \centering
     \includegraphics[width=\linewidth]{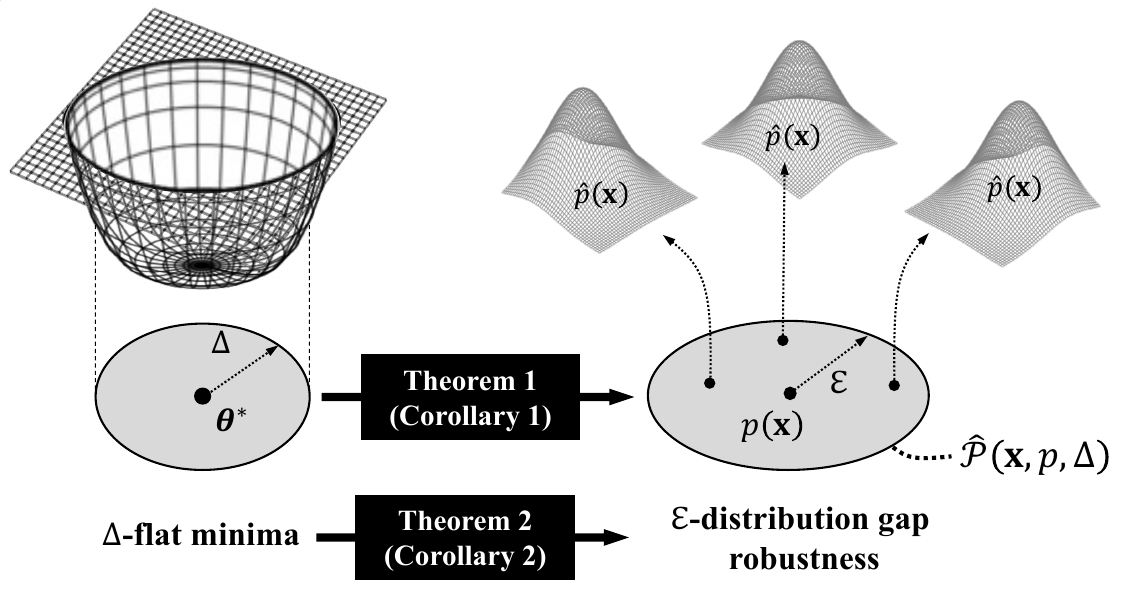}
     \vspace{-7mm}
     \caption{A conceptual illustration of theoretical analysis. \textbf{\cref{thm1}} (\textbf{\cref{cor1.1}} for diffusion model)  translates the perturbation in the parameter space into the set of perturbed distributions. \textbf{\cref{thm2}} (\textbf{\cref{cor2.1}} for diffusion model) shows that flat minima lead to robustness against the distribution gap.}
     \label{fig:thm}
     \vspace{-3mm}
 \end{figure}

\begin{table*}[t!]
\resizebox{\linewidth}{!}{
    \centering
    \renewcommand{\arraystretch}{1.3} 
    \setlength{\tabcolsep}{8pt}  
    \begin{tabular}{c c cc cccc}
        \toprule
        \multirow{2}{*}{\textbf{FID Score}} & \textbf{Dataset} & \multicolumn{2}{c}{\textbf{CIFAR-10 (32x32)}} & \multicolumn{2}{c}{\textbf{LSUN Tower (64x64)}} & \multicolumn{2}{c}{\textbf{FFHQ (64x64)}} \\ 
        \hhline{|~|-|-|-|-|-|-|-|}
        & \textbf{$T'$}
        & 20 steps 
        & 100 steps
        & 20 steps
        & 100 steps
        & 20 steps
        & 100 steps \\ \hline 
        \multirow{9}{*}{\textbf{Algorithms}} 
        &  ADM & 34.47 & 8.80 & 36.65 & 8.57 & 30.81 & 7.53  \\ 
        & \texttt{+EMA} & 10.63 & 4.06 & 7.87 & 2.49 & 19.03 & 6.19 \\
        & \texttt{+SWA} & 11.00 & 3.78 & 8.72 & 2.31 & 17.93 & 5.49 \\ 
        \hhline{|~|-|-|-|-|-|-|-|}
        & \texttt{+IP} & 20.11 & 7.23 &  25.77 & 7.00 & 15.03 & 13.55 \\ 
        & \texttt{+IP+EMA} & 9.10 & 3.46 & 7.66 & 2.43 & 11.72 & 4.00 \\ 
        & \texttt{+IP+SWA} & 9.04 & 3.07 & 8.55 & 2.34 & 12.99 & \textbf{3.54} \\ 
        \hhline{|~|-|-|-|-|-|-|-|}
        & \textbf{\texttt{+SAM}} & 9.01 & 3.83 & 16.02 & 4.79& 11.59 & 5.29  \\ 
        & \textbf{\texttt{+SAM+EMA}} & \textbf{7.00} & 3.18 & 6.66 & 2.30 & \textbf{11.41} & 5.04 \\ 
        & \textbf{\texttt{+SAM+SWA}} & 7.27 & \textbf{2.96} & \textbf{6.50} & \textbf{2.27} & 12.15 & 4.17 \\
        \bottomrule
    \end{tabular}
}
    \caption{
    FID Scores for ADM baselines with different algorithms on CIRAR-10, LSUN Tower, and FFHQ datasets. We use DDPM sampling with shorter respacing timesteps, $T'=20, 100$.}
    \label{tab:results}
    \vspace{-3mm}
\end{table*}

 \begin{theorem}\label{thm2}
(A link from $\Delta$-flatness to $\mathcal{E}$-gap robustness)
A $\Delta$-flat minimum achieves $\mathcal{E}$-distribution gap robustness, such that $\mathcal{E}$ is upper-bounded as follows:
\begin{gather}
\mathcal{E}\leq \max_{\hat{p}\sim\hat{\mathcal{P}}(\mathbf{x};p,\Delta)}D(p||\hat{p}).
\end{gather}
\end{theorem}
Let us further manipulate the upper bound for the diffusion model case. From \textbf{\cref{cor1.1}}, $\hat{p}$ is a form of Gaussian distribution; thus, it is possible to achieve the closed form of the upper bound as follows:
\begin{corollary}\label{cor2.1}
(Diffusion version of \textbf{\cref{thm2}})
For a diffusion model, a $\Delta$-flat minimum achieves $\mathcal{E}$-distribution gap robustness, such that $\mathcal{E}$ is upper-bounded as follows:
\begin{align}
\mathcal{E} &\leq \max_{\|\delta\|_2\leq \Delta} \frac{1}{2} \Bigg[ \log|\Sigma_{\delta}| -d + tr(\Sigma_{\delta}^{-1}) + {\boldsymbol\mu_{\delta}}^\top  {\Sigma_{\delta}}^{-1}\boldsymbol\mu_{\delta}
\Bigg] \nonumber\\
&\leq \frac{1}{2}\Bigg[ \sum_{i=1}^d (\sigma_i-\log \sigma_i) -d+
\frac{\sigma_{d}}{m^2}\|\mathbf{U}^\top\mathbf{e}\|_2^2\Delta^{2} 
\Bigg],
\end{align}
where $\sigma_i$ is an eigenvalue of $\Sigma_{\delta}^{-1}$ with the increasing order of $\sigma_{1} \,\le\, \sigma_{2} \,\le\, \dots \,\le\, \sigma_{d}$.
\end{corollary}

\begin{remark}\label{remark2.1}
\textbf{(Flatter minima enhance distribution gap robustness)} It is noteworthy that flatter minima with a large $\Delta$ flatten the loss values for the far-pushed away perturbed distribution $\hat{p}$ with large divergence, leading to the larger bound of $\mathcal{E}$. It indicates that flattening the loss surface makes the generative model zero-force the loss values of diversified prior distribution, thus enhancing the robustness against the prior distribution gap.
\end{remark}

\smallskip
\noindent
\textbf{Flatness reduces exposure bias (a gap of $\hat{\epsilon}$ and $\epsilon$):}
A clear advantage of flatness is robustness to exposure bias, where the errors in noise estimation accumulate over iterations, thus severely deteriorating the generative performance. In formula, $\hat{\epsilon}$ deviates from $\epsilon$. We argue that a generative model with a flat minimum suppresses the loss values of perturbed estimation, so that the error accumulation is sufficiently relieved.
To be shown in experiments, we empirically confirm that flat minima tend to show a smaller exposure bias, leading to robust generative performance.

\smallskip
\noindent
\textbf{Flatness becomes robust to model quantization (a compression from $\boldsymbol{\theta}$ to $\hat{\boldsymbol{\theta}}$):} For another benefit, flat minima make the model robust to the performance degradation caused by the model quantization.
Recently, the quantization of generative models has become crucial, when enabling real-time applications of generative tasks~\cite{Diffquant_ptq4dit, Diffquant_qdiffusion, Diffquant_brecq}. We claim that flat minima inherently bolster the robustness against the degradation due to the quantization, because the quantized model, $\hat{\boldsymbol{\theta}}$, can be viewed as a perturbed version of the full-precision parameter, i.e., $\hat{\boldsymbol{\theta}}=\boldsymbol{\theta}+\Delta$; thus the loss remains flat for the perturbation~\cite{Diffquant_brecq}.
In experiments, we clearly demonstrate that the diffusion model with flatness is remarkably resilient to degradation from quantization compared to other methods.

A conceptual overview of our theoretical analysis is illustrated in \cref{fig:thm}.
The proof of \textbf{\cref{thm1}}, 
and \textbf{\cref{cor1.1}}, \textbf{\ref{cor2.1}} are given in \Aref{appendix:proofs} in Supplementary.

\if false
\begin{theorem}
(Flat minima links to exposure bias robustness) If $\boldsymbol{\theta}^*$ is a $\Delta$-flat minimum, then the SGM $s_{\boldsymbol{\theta}}$ is $\mathcal{E}$-exposure bias robust such that,
\begin{equation}
\mathcal{E} \leq \argmax_{\|\delta\|_2\leq \Delta} \frac{1}{2} \Bigg[ \log|\Sigma_{\delta}| -d + tr(\Sigma_{\delta}^{-1}) + {\boldsymbol\mu_{\delta}}^\top  {\Sigma_{\delta}}^{-1}\boldsymbol\mu_{\delta}
\Bigg], \nonumber
\end{equation}
where $\Sigma_{\delta}:= I + \displaystyle\frac{\delta_w}{m}$, $\boldsymbol \mu_{\delta}:=\displaystyle\frac{1}{m}\Sigma_{\delta}$, $\delta_w:= \delta\mathbf{W}^\top$, and  $\delta_u:= \delta \mathbf{U}^\top\mathbf{e}$.
\end{theorem}

Exposure bias is the error caused by the discrepancy between $p(\mathbf{x})$ and $\hat{p}(\mathbf{x})$, and we define it as $D_{KL}(\hat{p}\| p)$, i.e., KL divergence.
By following the Theorem.\ref{thm: Eps-flat minima and Eps-perturbed distribution}, $\theta^*$ has flat loss values in $\mathcal{E}$-perturbed distribution, leading to robustness in distribution shift caused by $\epsilon$ model perturbation.

When we train the diffusion model, we add the noise $\boldsymbol \epsilon_{t-1}$ in the forward process and want the diffusion model to predict the $\boldsymbol \epsilon_{t-1}$ in the reverse process where $\boldsymbol \epsilon_{t-1}$ follows the normal Gaussian distribution, i.e., $\epsilon_{t}\sim \mathcal{N}(0, I) \: \forall \: t\in [T]$.
Therefore, the distribution that the model trains is the normal Gaussian, and we can denote the perturbed distribution in Def. \ref{Def: Set of Perturbed Distribution} as follows:
\fi

\section{Experiments}
\label{sec:Experiments}
\begin{figure*}
    \centering
    \includegraphics[width=\linewidth]{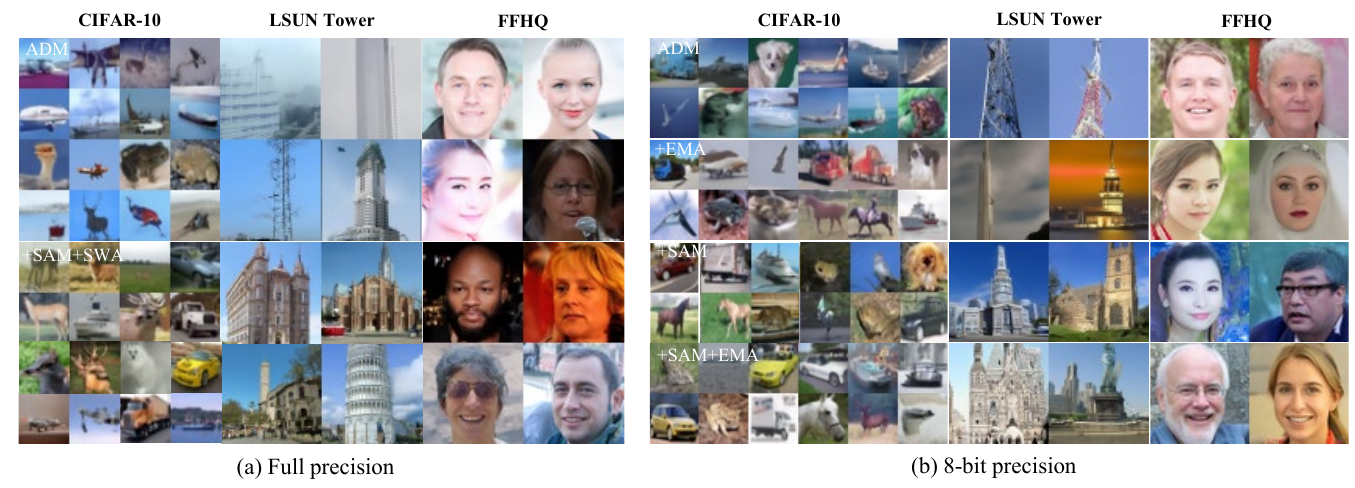}
    \vspace{-8mm} 
    \caption{Generation results of (a) full precision (32-bit) and (b) 8-bit quantized models. We use respaced timesteps $T'=20$ for sampling. For the 8-bit case, we selectively compare +EMA, +SAM, and their combinations, where the relatively flatter minima have been found.}
    \label{fig:generated_samples}
    \vspace{-5mm}
\end{figure*}
\subsection{Experiments Settings}

\noindent\textbf{Datasets.} Similar to the setup in~\cite{DDPM-IP}, we use CIFAR-10 (32x32), LSUN-tower (64x64), and FFHQ (64x64).

\smallskip
\noindent\textbf{Baselines.} We use ADM baselines\footnote{https://github.com/openai/guided-diffusion}~\cite{ADM} for training unconditional DDPM.  We incorporate flatness-enhancing approaches such as SWA~\cite{SAM'21} and SAM~\cite{SAM'21} to deliberately enhance flatness and reveal the effects of that. 
Because EMA may influence flatness~\cite{EMA}, we distinguish between baselines with and without EMA. 
Additionally, we compare DDPM-IP~\cite{DDPM-IP} that introduces input perturbation during training to encourage smoothness in the diffusion model.

\smallskip
\noindent\textbf{Experimental details.}
We trained the model for 200K steps on the CIFAR-10, LSUN Tower, and FFHQ datasets. We employed the Adam optimizer for all experiments with a learning rate of $1e^{-4}$, following ~\cite{ADM, ADM-ES}. For \texttt{+IP}, we used an input perturbation strength of 0.1 following~\cite{DDPM-IP}, and we set the cycle length $c$ to 100 and start averaging from 180K steps for \texttt{SWA}. Finally, we tuned the $\rho \in [1e^{-1}, 1e^{-2},1e^{-3}]$ parameters for \texttt{+SAM}.

\smallskip
\noindent\textbf{Evaluation metrics.} We evaluate generative performance by comparing Fr\'{e}chet Inception Distance (FID) scores~\cite{FID} using subsequence of timesteps $(T' < T)$. We assess Low-Pass Filter (LPF)~\cite{LPF-SGD'22} values and loss plots under perturbation~\cite{SWAD'21} to determine whether the baselines successfully identify a flat loss surface. To further investigate generalization ability of diffusion models, we investigate exposure bias and quantization error. For exposure bias~\cite{DDPM-ES}, we compare the square norm of the predicted noise, $\left\| \epsilon_{\theta} \right\|^2$ when models are conditioned on ground truth noisy images during training versus when they receive error-contained inputs at different sampling timesteps~\cite{ADM-ES}. 
For quantization error, we compare the FID before and after quantization.

\subsection{Generative Performance}
\begin{table}[t!]
\centering
\begin{tabular}{c | c c c }
\toprule
            LPF $\downarrow$ & \texttt{w/o} & \texttt{+EMA} & \texttt{+SWA}\\ \hline
            ADM & 0.097& 0.099& 0.099 \\
            \texttt{+IP} & 0.103& 0.101& 0.102 \\
            \texttt{+SAM} & \textbf{0.063} & \textbf{0.063}& \textbf{0.063} \\
\bottomrule
		\end{tabular}
    \footnotesize{
		\\ $\downarrow$: a lower value is preferred.}
\vspace{-3mm}
\caption{Flatness measure on CIFAR-10. We calculate the loss with the perturbed model with Gaussian noise.
Lower values indicate a flatter loss landscape.
}
\label{tab: flatness LPF}
\vspace{-4mm}
\end{table}
The quantitative and qualitative results presented in \cref{tab:results,fig:generated_samples} demonstrate the generative performance of baselines. 
From \cref{tab:results}, we observe that applying \texttt{EMA} consistently improves FID scores across all datasets and respaced timesteps, demonstrating its effectiveness in stabilizing model updates and enhancing sample quality. \texttt{SWA} also provides notable improvements but does not outperform \texttt{EMA} in most cases.\footnote{Because of limited space, we include the discussion and experimental results of the post-hoc EMA (pEMA)~\cite{rebuttal_EDM2} in \Aref{appendix:posthoc_EMA}}.
While EMA and SWA can always be applied as auxiliary options, both standalone \texttt{SAM} and \textbf{\texttt{+SAM+SWA}} achieve comparable or better FID scores, especially under 20 timesteps, where exposure bias worsens. This suggests that explicit flatness control improves generalization, in terms of enhancing sample quality and robustness.
For \texttt{IP}, the FID score is comparable to that of \texttt{SAM} at 100 steps, since the perturbation in the data distribution translates into parameter perturbations (\textbf{Theorem 1}). However, \texttt{IP} performs poorly at 20 steps, as it lacks a principled noise direction and tends to converge to suboptimal minima.
\Cref{fig:generated_samples} further supports these findings by providing visual comparisons of generated images
\footnote{We include additional sampling images in \Aref{appendix:fixed_seed}}.

\subsection{Flatness Measurement}
\begin{figure}[t!]
    \centering
    \includegraphics[width=\linewidth]{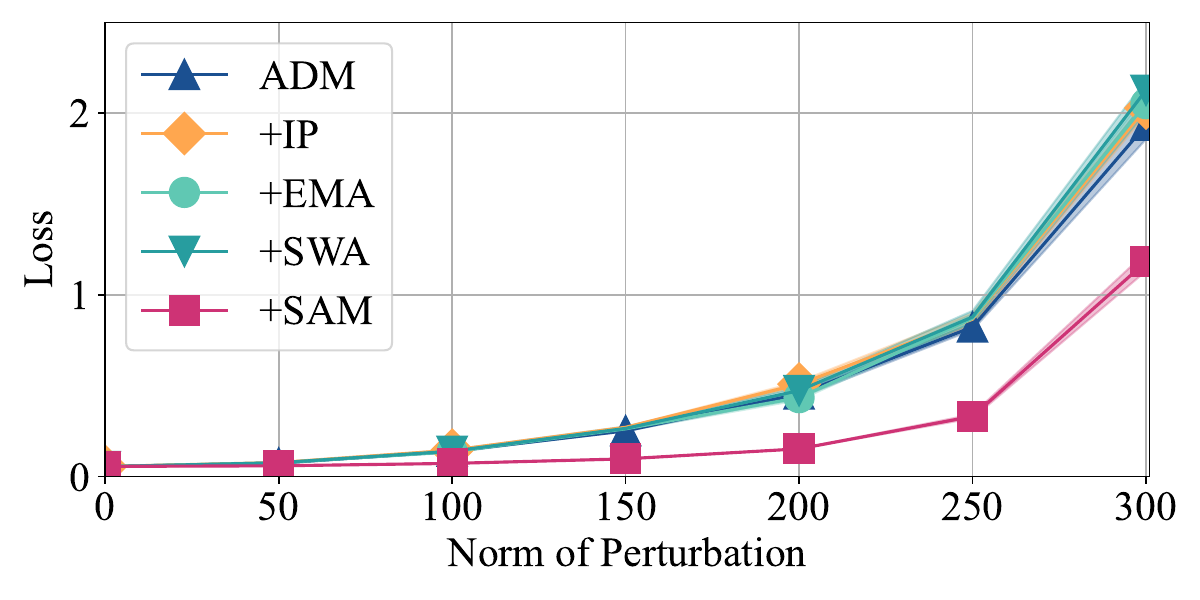}
    \vspace{-8mm} 
    \caption{
    Loss plots under perturbation for CIFAR-10. As the norm of model perturbation increases, we plot the corresponding loss values. A smaller slope indicates a flatter loss landscape.}
    \label{fig:flatness MC}
     \vspace{-5mm} 
\end{figure}

In \cref{tab: flatness LPF}, we compare LPF flatness measures~\cite{LPF-SGD'22}, where lower values indicate a flatter loss landscape. We observe that ADM already possesses a certain level of flatness. Interestingly, we find that ensemble-like averaging, such as \texttt{+SWA} and \texttt{+EMA}, fail to induce additional flatness. In contrast, \texttt{+SAM} finds a flatter loss surface by explicitly perturbing model weights, as evidenced by \texttt{SAM}-applied models consistently achieving the lowest LPF values and significantly superior performance to other baselines. Also, we plot another measurement to support these findings further. \Cref{fig:flatness MC} shows how the loss value increases as the model perturbation is imposed for CIFAR-10. 
\texttt{+IP}, \texttt{+EMA}, and \texttt{+SWA} show no significant differences, coincide with the result of LPF, where these fail to find additional flatness. In this setting, \texttt{+SAM} shows a flatter behavior than other algorithms. It shows that diffusion models already show flatter loss landscapes, and ensemble-like averaging, such as \texttt{+SWA} and \texttt{EMA}, shows less impact. Moreover, finding flat minima explicitly leads to the flatness in diffusion models. We provide plots for other algorithms, including \texttt{+IP+EMA}, \texttt{+IP+SWA}, \texttt{+SAM+EMA}, \texttt{+SAM+SWA}, in the \Aref{appendix:further_results}.

\subsection{Flat Diffusion Models are Robust}
\begin{figure}[t!]
    \centering
    \includegraphics[width=\linewidth]{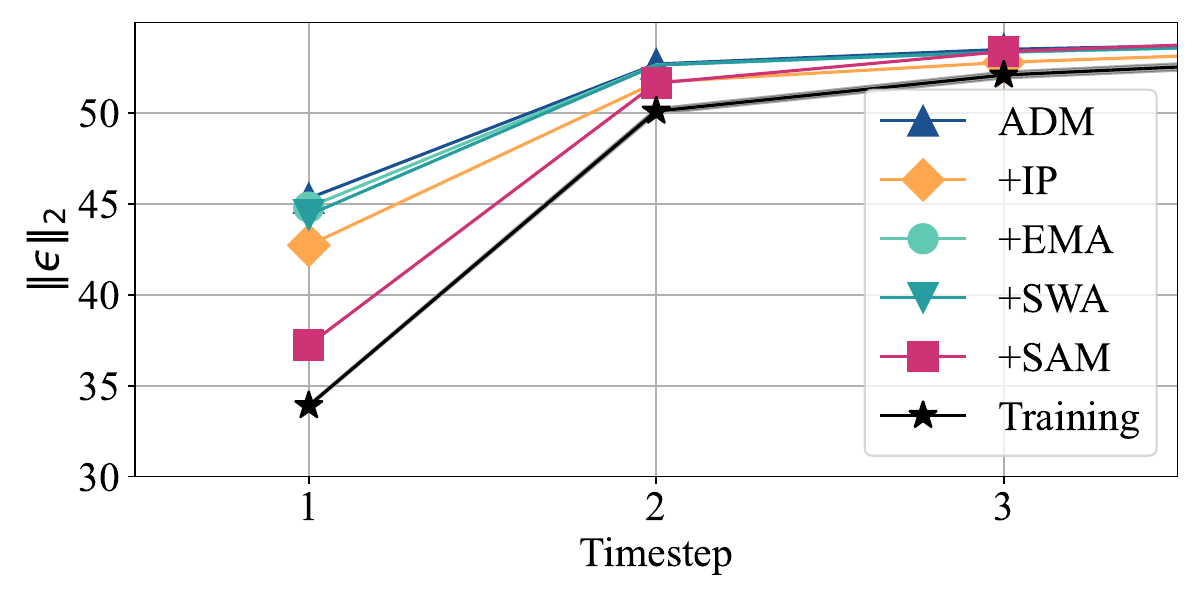}
    \vspace{-8mm} 
    \caption{L2 norm of the predicted noise for CIFAR-10. As the gap with ``Training'' decreases, the predicted noise norm during sampling approaches the ground truth.}
    \label{fig:eps norm} 
    \vspace{-5mm} 
\end{figure}
We show that flat diffusion models perform well under input shifts. As an instantiation, we set two cases: 1) exposure bias and 2) quantization error. Both scenarios share the common characteristic that diffusion models receive error-accumulated inputs, leading to performance degradation.

\smallskip
\noindent
\textbf{Exposure bias.} 
In \cref{fig:eps norm}, we compare the L2 norm of the predicted noise in diffusion models. A model that closely aligns with the training trajectory is expected to exhibit lower exposure bias, thereby mitigating distribution shifts over iterative sampling steps. We observe that \texttt{SAM}-applied models consistently maintain L2 norms closer to the training curve, indicating a lower degree of exposure bias. This result aligns well with their superior FID performance reported in ~\cref{tab:results}, suggesting that flat diffusion models generalize better across sampling timesteps and effectively combat error accumulation.
\begin{table}[t!]
\centering
\begin{tabular}{c | c | c }
\toprule
 32-bit $\rightarrow$ 8-bit & 20 steps & 100 steps \\
\midrule
 ADM & 34.47 $\overset{\text{{\tiny\textcolor{red}{+13.65}}}}{\longrightarrow}$ 48.02 & 8.80 $\overset{\text{{\tiny\textcolor{red}{+3.98}}}}{\longrightarrow}$ 12.78  \\
 \texttt{+EMA} & 10.63 $\overset{\text{{\tiny\textcolor{red}{+10.02}}}}{\longrightarrow}$ 20.65 & 4.06 $\overset{\text{{\tiny\textcolor{red}{+3.3}}}}{\longrightarrow}$ 7.36 \\
 \texttt{+SAM} & 9.01 $\overset{\text{{\tiny\textcolor{green!60!black}{-0.07}}}}{\longrightarrow}$ 8.94 & 3.83 $\overset{\text{{\tiny\textcolor{red}{+0.19}}}}{\longrightarrow}$ 4.02 \\
 \texttt{+SAM+EMA} & 7.00 $\overset{\text{{\tiny\textcolor{red}{+0.2}}}}{\longrightarrow}$ 7.20 & 3.18 $\overset{\text{{\tiny\textcolor{green!60!black}{-0.06}}}}{\longrightarrow}$ 3.12  \\
\bottomrule
\end{tabular}
\vspace{-3mm} 
\caption{FID performance on CIFAR-10 when model parameters are quantized to 8-bit. We apply direct quantization by clipping and rounding 32-bit parameters without additional fine-tuning. The 32-bit performance results are identical to those in ~\Cref{tab:results}.}
\vspace{-4mm}
\label{tab:quantization}
\end{table}

\smallskip
\noindent
\textbf{Model quantization error.} \cref{tab:quantization} presents the FID scores of quantization across baselines, when reducing precision from 32-bit to 8-bit. Here, we quantize model parameters to 8-bit values using scaling and rounding, without any additional fine-tuning or retraining.
Thanks to \texttt{+SAM} that aims to the robustness of model perturbation, flatter diffusion models results in surprisingly better robustness under quantization.
These findings also support that \texttt{+SAM} is robust to model quantization. 

\section{Related works}
\label{sec:Related works}
\textbf{Diffusion models.}
Diffusion probabilistic models (DPMs)~\cite{DDPM, ADM, SGM1, SGM2, SGM_SDE} have recently emerged as powerful generative frameworks, achieving state-of-the-art performance in image~\cite{LDM, DiT} and video~\cite{HVDM, StableVDM} synthesis. Despite extensive research on improving diffusion models through architecture modifications~\cite{iDDPM, EDM, DiT}, enhanced training strategies~\cite{PatchDiffusion}, and novel noise scheduling techniques~\cite{SGM_SDE, ConsistencyModel}, relatively little attention has been given to understanding their loss landscape properties.

\smallskip
\noindent
\textbf{Flat minima in various tasks.}
Flat minima have garnered significant attention in classification and domain generalization tasks with the view of loss landscape ~\cite{SWA'18,Loss-surface'18,Loss-landscape'18} and objective function~\cite{SAM'21,ASAM'21}. Other tasks also adopt the flatness for their performance. In federated learning, for example, where client devices collect and train on heterogeneous datasets, ensuring robust generalization across diverse data distributions is a core challenge. 
Recent works have employed flat minima strategies to mitigate this data heterogeneity issue by obtaining the model robust to distribution shift~\cite{FedSAM'22-Theory,FedSAM'22-Experiment,FedGF} and enhance the robustness of compressed neural networks~\cite{flatness_compression_1, flatness_compression_2}.

\smallskip
\noindent\textbf{Flat minima on diffusion models.}
Recent findings suggest that diffusion models inherently exhibit a flatter loss landscape~\cite{WhyDiffusionFlat}. However, they do not elucidate why the landscape is flat or how it relates to generalization and robustness, leaving the role of flatness in diffusion models unexplored.
While prior works~\cite{WGAN, GS-WGAN, DDPM-IP} leverage smoothness via Lipschitz continuity or gradient regularization, they do not directly examine flatness. To the best of our knowledge, we are the first to systematically investigate how flat minima influence diffusion models, their generative performance, and robustness--motivating our theoretical and empirical analysis.

\section{Discussion and Limitation}
\label{sec:discussion}

The generative modeling community has long sought to understand generalization in terms of preventing models from overfitting to the training set while ensuring the generation of diverse and high-quality samples, particularly for underrepresented classes~\cite{Memorization_class_1, Memorization_class_2}. 
While our work primarily investigates the role of flat minima in addressing exposure bias and quantization error, 
\begin{wraptable}{r}{0.4\linewidth}
  \centering
  \vspace{-3mm}
  \begin{tabular}{cc}
   & \footnotesize{FID $\downarrow$} \\
    \hline
    \footnotesize{ADM} & \footnotesize{15.97 (+5.34)} \\    \rowcolor{SkyBlue!30!white}\footnotesize{\textbf{\texttt{+SAM}}} & \footnotesize{\textbf{11.72} (\textbf{+4.72})} \\
    \makecell{
    \footnotesize{\textbf{\texttt{+SAM}}}\vspace{-3pt} \\ \tiny{(7$\times$ noise)}
    }
    &  \footnotesize{15.72 (+8.72)} \\
    \hline
  \end{tabular}
  \vspace{-1.5mm}
  \caption{FID degradations under adversarial attack.}
  \label{tab:adv_attack}
  \vspace{-3mm}
\end{wraptable}
We conjecture that flat diffusion models may offer broader implications, such as cross-domain generation capability, few-shot personalization, outlier sample generation, mitigating memorization of generative models, and adversarial attack.
In \cref{tab:adv_attack}, we simulate an adversarial attack by maximizing the diffusion loss w.r.t. $\tilde{z}_T$. \textbf{\texttt{+SAM}} consistently achieves better FID than ADM and maintains its performance even under 7$\times$ amplified adversarial noise.
Given the link between flatness and generalization, this insight suggests promising directions for future research into their role in generative modeling. As an initial attempt, we validate the benefits of flatness on relatively simple, low-resolution tasks. Notably, our theoretical claim is not confined to data dimensionality; extending the analysis to more complex settings would further support our theoretical claims.

\section{Conclusion}
\label{sec:conclusion}
We presented an in-depth investigation of the role and benefits of flat loss surfaces in generative models, which have remained unknown despite the great success of deep generative models.
Our study explored this unknown by analyzing how flatness influences generative modeling and demonstrating its possible impact on robustness. 
First, our theoretical analysis reveals that the loss flatness of the generative models enhances the robustness against the disruptions of the target data distributions or the perturbations of model parameters. 
These insights naturally link to the reduced exposure bias problem and the model quantization error.
Based on the evaluations on CIFAR-10, LSUN Tower, and FFHQ datasets, we demonstrated that 1) SAM is suitable to seek flatter minima of diffusion models rather than SWA and EMA, 2) flat minima reduce exposure bias with a minimal gap between testing and training, and 3) a flat model keeps its generative performance even with a strong model quantization; coinciding with our theory.

\section*{Acknowledgments}
This work was supported by the National Research Foundation of Korea (NRF) grant funded by
the Korea government (MSIT) (No.2022R1C1C100849612, No.RS-2024-00459023), Institute of Information \& Communications Technology Planning \& Evaluation (IITP) grant funded by the Korea government (MSIT):(No.RS-2020-II201336, Artificial Intelligence Graduate School Program at UNIST), (RS-2025-25442824, AI Star Fellowship Program at UNIST), (No.2022-0-00959, No.RS-2022-II220959 (Part 2) Few-Shot Learning of Causal Inference in Vision and Language for Decision Making), (RS-2025-25441996, Development of a Virtual Tactile Signal Generation Platform Technology Based on Multimodal Vision-Tactile Integrated AGI), (RS-2025-25442149), Korea Health Technology R\&D Project through the Korea Health Industry Development Institute (KHIDI), funded by the Ministry of Health \& Welfare, Republic of Korea (No.RS-2025-02223382) and 2025 Research Fund (1.250006.01) of UNIST(Ulsan National Institute of Science \& Technology.

{
    \small
    \bibliographystyle{ieeenat_fullname}
    \bibliography{main}

\begin{thebibliography}{48}
\providecommand{\natexlab}[1]{#1}
\providecommand{\url}[1]{\texttt{#1}}
\expandafter\ifx\csname urlstyle\endcsname\relax
  \providecommand{\doi}[1]{doi: #1}\else
  \providecommand{\doi}{doi: \begingroup \urlstyle{rm}\Url}\fi

\bibitem[Arjovsky et~al.(2017)Arjovsky, Chintala, and Bottou]{WGAN}
Martin Arjovsky, Soumith Chintala, and L{\'e}on Bottou.
\newblock Wasserstein generative adversarial networks.
\newblock In \emph{International conference on machine learning}, pages
  214--223. PMLR, 2017.

\bibitem[Bisla et~al.(2022)Bisla, Wang, and Choromanska]{LPF-SGD'22}
Devansh Bisla, Jing Wang, and Anna Choromanska.
\newblock Low-pass filtering sgd for recovering flat optima in the deep
  learning optimization landscape.
\newblock In \emph{International Conference on Artificial Intelligence and
  Statistics}, pages 8299--8339. PMLR, 2022.

\bibitem[Blattmann et~al.(2023)Blattmann, Dockhorn, Kulal, Mendelevitch,
  Kilian, Lorenz, Levi, English, Voleti, Letts, et~al.]{StableVDM}
Andreas Blattmann, Tim Dockhorn, Sumith Kulal, Daniel Mendelevitch, Maciej
  Kilian, Dominik Lorenz, Yam Levi, Zion English, Vikram Voleti, Adam Letts,
  et~al.
\newblock Stable video diffusion: Scaling latent video diffusion models to
  large datasets.
\newblock \emph{arXiv preprint arXiv:2311.15127}, 2023.

\bibitem[Caldarola et~al.(2022)Caldarola, Caputo, and
  Ciccone]{FedSAM'22-Experiment}
Debora Caldarola, Barbara Caputo, and Marco Ciccone.
\newblock Improving generalization in federated learning by seeking flat
  minima.
\newblock In \emph{European Conference on Computer Vision (ECCV)}, pages
  654--672. Springer, 2022.

\bibitem[Cha et~al.(2021)Cha, Chun, Lee, Cho, Park, Lee, and Park]{SWAD'21}
Junbum Cha, Sanghyuk Chun, Kyungjae Lee, Han-Cheol Cho, Seunghyun Park, Yunsung
  Lee, and Sungrae Park.
\newblock Swad: Domain generalization by seeking flat minima.
\newblock \emph{Advances in Neural Information Processing Systems},
  34:\penalty0 22405--22418, 2021.

\bibitem[Chen et~al.(2020)Chen, Orekondy, and Fritz]{GS-WGAN}
Dingfan Chen, Tribhuvanesh Orekondy, and Mario Fritz.
\newblock Gs-wgan: A gradient-sanitized approach for learning differentially
  private generators.
\newblock \emph{Advances in Neural Information Processing Systems},
  33:\penalty0 12673--12684, 2020.

\bibitem[Dhariwal and Nichol(2021)]{ADM}
Prafulla Dhariwal and Alexander Nichol.
\newblock Diffusion models beat gans on image synthesis.
\newblock \emph{Advances in neural information processing systems},
  34:\penalty0 8780--8794, 2021.

\bibitem[et~al.(2024)]{rebuttal_EDM2}
Karras et al.
\newblock Analyzing and improving the training dynamics of diffusion models.
\newblock In \emph{CVPR}, 2024.

\bibitem[Foret et~al.(2021)Foret, Kleiner, Mobahi, and Neyshabur]{SAM'21}
Pierre Foret, Ariel Kleiner, Hossein Mobahi, and Behnam Neyshabur.
\newblock Sharpness-aware minimization for efficiently improving
  generalization.
\newblock In \emph{International Conference on Learning Representations}, 2021.

\bibitem[Garipov et~al.(2018)Garipov, Izmailov, Podoprikhin, Vetrov, and
  Wilson]{Loss-surface'18}
Timur Garipov, Pavel Izmailov, Dmitrii Podoprikhin, Dmitry~P Vetrov, and
  Andrew~G Wilson.
\newblock Loss surfaces, mode connectivity, and fast ensembling of dnns.
\newblock \emph{Advances in neural information processing systems}, 31, 2018.

\bibitem[Granziol et~al.(2020)Granziol, Wan, Albanie, and Roberts]{IterAvg'20}
Diego Granziol, Xingchen Wan, Samuel Albanie, and Stephen Roberts.
\newblock Iterative averaging in the quest for best test error.
\newblock \emph{arXiv preprint arXiv:2003.01247}, 2020.

\bibitem[Heusel et~al.(2017)Heusel, Ramsauer, Unterthiner, Nessler, and
  Hochreiter]{FID}
Martin Heusel, Hubert Ramsauer, Thomas Unterthiner, Bernhard Nessler, and Sepp
  Hochreiter.
\newblock Gans trained by a two time-scale update rule converge to a local nash
  equilibrium.
\newblock \emph{Advances in neural information processing systems}, 30, 2017.

\bibitem[Ho et~al.(2020)Ho, Jain, and Abbeel]{DDPM}
Jonathan Ho, Ajay Jain, and Pieter Abbeel.
\newblock Denoising diffusion probabilistic models.
\newblock \emph{Advances in neural information processing systems},
  33:\penalty0 6840--6851, 2020.

\bibitem[Izmailov et~al.(2018)Izmailov, Podoprikhin, Garipov, Vetrov, and
  Wilson]{SWA'18}
Pavel Izmailov, Dmitrii Podoprikhin, Timur Garipov, Dmitry Vetrov, and
  Andrew~Gordon Wilson.
\newblock Averaging weights leads to wider optima and better generalization.
\newblock \emph{arXiv preprint arXiv:1803.05407}, 2018.

\bibitem[Karras et~al.(2022)Karras, Aittala, Aila, and Laine]{EDM}
Tero Karras, Miika Aittala, Timo Aila, and Samuli Laine.
\newblock Elucidating the design space of diffusion-based generative models.
\newblock \emph{Advances in neural information processing systems},
  35:\penalty0 26565--26577, 2022.

\bibitem[Kim et~al.(2024)Kim, Lee, Park, Kim, Lee, Kim, and Yoo]{HVDM}
Kihong Kim, Haneol Lee, Jihye Park, Seyeon Kim, Kwanghee Lee, Seungryong Kim,
  and Jaejun Yoo.
\newblock Hybrid video diffusion models with 2d triplane and 3d wavelet
  representation.
\newblock In \emph{European Conference on Computer Vision}, pages 148--165.
  Springer, 2024.

\bibitem[Klinker(2011)]{EMA}
Frank Klinker.
\newblock Exponential moving average versus moving exponential average.
\newblock \emph{Mathematische Semesterberichte}, 58:\penalty0 97--107, 2011.

\bibitem[Kwon et~al.(2021)Kwon, Kim, Park, and Choi]{ASAM'21}
Jungmin Kwon, Jeongseop Kim, Hyunseo Park, and In~Kwon Choi.
\newblock Asam: Adaptive sharpness-aware minimization for scale-invariant
  learning of deep neural networks.
\newblock In \emph{International Conference on Machine Learning}, pages
  5905--5914. PMLR, 2021.

\bibitem[Lee and Yoon(2024)]{FedGF}
Taehwan Lee and Sung~Whan Yoon.
\newblock Rethinking the flat minima searching in federated learning.
\newblock In \emph{Forty-first International Conference on Machine Learning},
  2024.

\bibitem[Li et~al.(2018)Li, Xu, Taylor, Studer, and
  Goldstein]{Loss-landscape'18}
Hao Li, Zheng Xu, Gavin Taylor, Christoph Studer, and Tom Goldstein.
\newblock Visualizing the loss landscape of neural nets.
\newblock \emph{Advances in neural information processing systems}, 31, 2018.

\bibitem[Li et~al.(2023{\natexlab{a}})Li, Qu, Yao, Sun, and Moens]{DDPM-TS}
Mingxiao Li, Tingyu Qu, Ruicong Yao, Wei Sun, and Marie-Francine Moens.
\newblock Alleviating exposure bias in diffusion models through sampling with
  shifted time steps.
\newblock \emph{arXiv preprint arXiv:2305.15583}, 2023{\natexlab{a}}.

\bibitem[Li et~al.(2023{\natexlab{b}})Li, Li, Zhang, and
  Bian]{GeneralizationDiffModel}
Puheng Li, Zhong Li, Huishuai Zhang, and Jiang Bian.
\newblock On the generalization properties of diffusion models.
\newblock \emph{Advances in Neural Information Processing Systems},
  36:\penalty0 2097--2127, 2023{\natexlab{b}}.

\bibitem[Li et~al.(2024{\natexlab{a}})Li, Liu, Tian, Wang, Wang, Jin, Wu, Tan,
  Lin, Liu, et~al.]{SwitchEMA}
Siyuan Li, Zicheng Liu, Juanxi Tian, Ge Wang, Zedong Wang, Weiyang Jin, Di Wu,
  Cheng Tan, Tao Lin, Yang Liu, et~al.
\newblock Switch ema: A free lunch for better flatness and sharpness.
\newblock \emph{arXiv preprint arXiv:2402.09240}, 2024{\natexlab{a}}.

\bibitem[Li et~al.(2024{\natexlab{b}})Li, Zhou, He, Cheng, and Huang]{F-SAM'24}
Tao Li, Pan Zhou, Zhengbao He, Xinwen Cheng, and Xiaolin Huang.
\newblock Friendly sharpness-aware minimization.
\newblock In \emph{Proceedings of the IEEE/CVF Conference on Computer Vision
  and Pattern Recognition (CVPR)}, pages 5631--5640, 2024{\natexlab{b}}.

\bibitem[Li et~al.(2023{\natexlab{c}})Li, Liu, Lian, Yang, Dong, Kang, Zhang,
  and Keutzer]{Diffquant_qdiffusion}
Xiuyu Li, Yijiang Liu, Long Lian, Huanrui Yang, Zhen Dong, Daniel Kang,
  Shanghang Zhang, and Kurt Keutzer.
\newblock Q-diffusion: Quantizing diffusion models.
\newblock In \emph{Proceedings of the IEEE/CVF International Conference on
  Computer Vision}, pages 17535--17545, 2023{\natexlab{c}}.

\bibitem[Li et~al.(2021)Li, Gong, Tan, Yang, Hu, Zhang, Yu, Wang, and
  Gu]{Diffquant_brecq}
Yuhang Li, Ruihao Gong, Xu Tan, Yang Yang, Peng Hu, Qi Zhang, Fengwei Yu, Wei
  Wang, and Shi Gu.
\newblock Brecq: Pushing the limit of post-training quantization by block
  reconstruction.
\newblock \emph{arXiv preprint arXiv:2102.05426}, 2021.

\bibitem[Liu et~al.(2022)Liu, Mai, Cheng, Chen, Hsieh, and You]{R-SAM'22}
Yong Liu, Siqi Mai, Minhao Cheng, Xiangning Chen, Cho-Jui Hsieh, and Yang You.
\newblock Random sharpness-aware minimization.
\newblock In \emph{Advances in Neural Information Processing Systems}, pages
  24543--24556. Curran Associates, Inc., 2022.

\bibitem[Na et~al.(2022)Na, Mehta, and Strubell]{flatness_compression_1}
Clara Na, Sanket~Vaibhav Mehta, and Emma Strubell.
\newblock Train flat, then compress: Sharpness-aware minimization learns more
  compressible models.
\newblock \emph{arXiv preprint arXiv:2205.12694}, 2022.

\bibitem[Nichol and Dhariwal(2021)]{iDDPM}
Alexander~Quinn Nichol and Prafulla Dhariwal.
\newblock Improved denoising diffusion probabilistic models.
\newblock In \emph{International conference on machine learning}, pages
  8162--8171. PMLR, 2021.

\bibitem[Ning et~al.(2023{\natexlab{a}})Ning, Li, Su, Salah, and
  Ertugrul]{ADM-ES}
Mang Ning, Mingxiao Li, Jianlin Su, Albert~Ali Salah, and Itir~Onal Ertugrul.
\newblock Elucidating the exposure bias in diffusion models.
\newblock \emph{arXiv preprint arXiv:2308.15321}, 2023{\natexlab{a}}.

\bibitem[Ning et~al.(2023{\natexlab{b}})Ning, Li, Su, Salah, and
  Ertugrul]{DDPM-ES}
Mang Ning, Mingxiao Li, Jianlin Su, Albert~Ali Salah, and Itir~Onal Ertugrul.
\newblock Elucidating the exposure bias in diffusion models.
\newblock \emph{arXiv preprint arXiv:2308.15321}, 2023{\natexlab{b}}.

\bibitem[Ning et~al.(2023{\natexlab{c}})Ning, Sangineto, Porrello, Calderara,
  and Cucchiara]{DDPM-IP}
Mang Ning, Enver Sangineto, Angelo Porrello, Simone Calderara, and Rita
  Cucchiara.
\newblock Input perturbation reduces exposure bias in diffusion models.
\newblock \emph{arXiv preprint arXiv:2301.11706}, 2023{\natexlab{c}}.

\bibitem[Peebles and Xie(2023)]{DiT}
William Peebles and Saining Xie.
\newblock Scalable diffusion models with transformers.
\newblock In \emph{Proceedings of the IEEE/CVF international conference on
  computer vision}, pages 4195--4205, 2023.

\bibitem[Qu et~al.(2022)Qu, Li, Duan, Liu, Tang, and Lu]{FedSAM'22-Theory}
Zhe Qu, Xingyu Li, Rui Duan, Yao Liu, Bo Tang, and Zhuo Lu.
\newblock Generalized federated learning via sharpness aware minimization.
\newblock In \emph{International Conference on Machine Learning (ICML)}, pages
  18250--18280. PMLR, 2022.

\bibitem[Ramachandran et~al.(2024)Ramachandran, Mukhopadhyay, Agarwal, Jawahar,
  and Namboodiri]{Memorization_class_1}
Sai~Niranjan Ramachandran, Rudrabha Mukhopadhyay, Madhav Agarwal, CV Jawahar,
  and Vinay Namboodiri.
\newblock Understanding the generalization of pretrained diffusion models on
  out-of-distribution data.
\newblock In \emph{Proceedings of the AAAI Conference on Artificial
  Intelligence}, pages 14767--14775, 2024.

\bibitem[Rombach et~al.(2022)Rombach, Blattmann, Lorenz, Esser, and Ommer]{LDM}
Robin Rombach, Andreas Blattmann, Dominik Lorenz, Patrick Esser, and Bj{\"o}rn
  Ommer.
\newblock High-resolution image synthesis with latent diffusion models.
\newblock In \emph{Proceedings of the IEEE/CVF conference on computer vision
  and pattern recognition}, pages 10684--10695, 2022.

\bibitem[SHI et~al.(2021)SHI, CHEN, Zhang, Zhan, and Wu]{OvercomingForget'21}
Guangyuan SHI, JIAXIN CHEN, Wenlong Zhang, Li-Ming Zhan, and Xiao-Ming Wu.
\newblock Overcoming catastrophic forgetting in incremental few-shot learning
  by finding flat minima.
\newblock In \emph{Advances in Neural Information Processing Systems}, pages
  6747--6761. Curran Associates, Inc., 2021.

\bibitem[Sohl-Dickstein et~al.(2015)Sohl-Dickstein, Weiss, Maheswaranathan, and
  Ganguli]{DDPM_nonequilibrium_thermodynamics}
Jascha Sohl-Dickstein, Eric Weiss, Niru Maheswaranathan, and Surya Ganguli.
\newblock Deep unsupervised learning using nonequilibrium thermodynamics.
\newblock In \emph{International conference on machine learning}, pages
  2256--2265. pmlr, 2015.

\bibitem[Song and Ermon(2019)]{SGM1}
Yang Song and Stefano Ermon.
\newblock Generative modeling by estimating gradients of the data distribution.
\newblock \emph{Advances in neural information processing systems}, 32, 2019.

\bibitem[Song and Ermon(2020)]{SGM2}
Yang Song and Stefano Ermon.
\newblock Improved techniques for training score-based generative models.
\newblock \emph{Advances in neural information processing systems},
  33:\penalty0 12438--12448, 2020.

\bibitem[Song et~al.(2020{\natexlab{a}})Song, Garg, Shi, and Ermon]{SGM3}
Yang Song, Sahaj Garg, Jiaxin Shi, and Stefano Ermon.
\newblock Sliced score matching: A scalable approach to density and score
  estimation.
\newblock In \emph{Uncertainty in artificial intelligence}, pages 574--584.
  PMLR, 2020{\natexlab{a}}.

\bibitem[Song et~al.(2020{\natexlab{b}})Song, Sohl-Dickstein, Kingma, Kumar,
  Ermon, and Poole]{SGM_SDE}
Yang Song, Jascha Sohl-Dickstein, Diederik~P Kingma, Abhishek Kumar, Stefano
  Ermon, and Ben Poole.
\newblock Score-based generative modeling through stochastic differential
  equations.
\newblock \emph{arXiv preprint arXiv:2011.13456}, 2020{\natexlab{b}}.

\bibitem[Song et~al.(2023)Song, Dhariwal, Chen, and
  Sutskever]{ConsistencyModel}
Yang Song, Prafulla Dhariwal, Mark Chen, and Ilya Sutskever.
\newblock Consistency models.
\newblock 2023.

\bibitem[Wang et~al.(2023)Wang, Jiang, Zheng, Wang, He, Wang, Chen, Zhou,
  et~al.]{PatchDiffusion}
Zhendong Wang, Yifan Jiang, Huangjie Zheng, Peihao Wang, Pengcheng He,
  Zhangyang Wang, Weizhu Chen, Mingyuan Zhou, et~al.
\newblock Patch diffusion: Faster and more data-efficient training of diffusion
  models.
\newblock \emph{Advances in neural information processing systems},
  36:\penalty0 72137--72154, 2023.

\bibitem[Wu et~al.(2025)Wu, Wang, Shang, Shah, and Yan]{Diffquant_ptq4dit}
Junyi Wu, Haoxuan Wang, Yuzhang Shang, Mubarak Shah, and Yan Yan.
\newblock Ptq4dit: Post-training quantization for diffusion transformers.
\newblock \emph{Advances in Neural Information Processing Systems},
  37:\penalty0 62732--62755, 2025.

\bibitem[Xu et~al.()Xu, Mi, Wang, and Chen]{WhyDiffusionFlat}
Tianshuo Xu, Peng Mi, Ruilin Wang, and Yingcong Chen.
\newblock Why diffusion models are stable and how to make them faster: An
  empirical investigation and optimization.

\bibitem[Yoon et~al.(2023)Yoon, Choi, Kwon, and Ryu]{Memorization_class_2}
TaeHo Yoon, Joo~Young Choi, Sehyun Kwon, and Ernest~K Ryu.
\newblock Diffusion probabilistic models generalize when they fail to memorize.
\newblock In \emph{ICML 2023 workshop on structured probabilistic inference
  $\{$$\backslash$\&$\}$ generative modeling}, 2023.

\bibitem[Zhou et~al.(2023)Zhou, Yang, Chang, and
  Mahoney]{flatness_compression_2}
Yefan Zhou, Yaoqing Yang, Arin Chang, and Michael~W Mahoney.
\newblock A three-regime model of network pruning.
\newblock In \emph{International Conference on Machine Learning}, pages
  42790--42809. PMLR, 2023.

\end{thebibliography}
}

\clearpage
\appendix
\onecolumn

\section*{Appendix}
\label{sec:supplementary}

\section{Mathematical claims and proofs}
\label{appendix:proofs}
For the main claims, we follow $\mathcal{L}(\mathbf{x};\boldsymbol{\theta},t,p_t) := ||s_{\boldsymbol{\theta}}(\mathbf{x},t)-\nabla_{\mathbf{x}}\log{p_t(\mathbf{x})}||_{2}^{2}$, while dropping the timestep $t$ without loss of generality. Our mathematical claims are valid for all timesteps.

\textbf{Definition 1.}
\textit{($\Delta$-flat minima)
 Let us consider a SGM with loss function $\mathcal{L}(\mathbf{x};\boldsymbol{\theta},p)$. A minimum $\boldsymbol{\theta^{*}}$ is $\Delta$-flat minima when the following constraints are hold:
\begin{gather}
\forall \: \delta\in\mathbb{R}^{d\times m} \text{ s.t. } \|\delta\|_{2}\leq\Delta, \:\: \mathcal{L}(\mathbf{x};\boldsymbol{\theta}^{*}+\delta,p) = l^* \nonumber\\
\exists \: \delta\in\mathbb{R}^{d\times m}
\text{ s.t. } \|\delta\|_{2}>\Delta, \:\:
\mathcal{L}(\mathbf{x};\boldsymbol{\theta}^*+\delta,p) > l^*, \nonumber
\end{gather}
where $l^*:=\mathcal{L}(\mathbf{x};\boldsymbol{\theta}^*,p)$ and $\Delta\in\mathbb{R}^{+}$.\footnote{$\forall$ means `for all,' $\exists$ means `there exists,' and $\mathbb{R}^+$ indicates the set of positive real numbers}
}

\textbf{Definition 2.}
\textit{($\mathcal{E}$-distribution gap robustness) 
A minimum $\boldsymbol{\theta^{*}}$ is $\mathcal{E}$-distribution gap robust when the following constraints are hold:
\begin{gather}
\forall \: \hat{p}(\mathbf{x}) \text{ s.t. } D(p||\hat{p}) \leq \mathcal{E}, \:\: \mathcal{L}(\mathbf{x};\boldsymbol{\theta}^{*},\hat{p}) = l^* \nonumber\\
\exists \: \hat{p}(\mathbf{x}) \text{ s.t. } D(p||\hat{p}) > \mathcal{E}, \:\: \mathcal{L}(\mathbf{x};\boldsymbol{\theta}^{*},\hat{p}) > l^*, \nonumber
\end{gather}
where $D(\cdot||\cdot)$ is the divergence between two probability density functions, $\hat{p}$ is the perturbed prior distribution of $\mathbf{x}$, and $\mathcal{E}$ is a positive real number.
}

\textbf{Theorem 1.}\textit{
(A perturbed distribution)
For a given prior distribution of $p(\mathbf{x})$ and the $\delta$-perturbed minimum, i.e., $\boldsymbol{\theta}+\delta$, the following $\hat{p}(\mathbf{x})$ satisfies the $\mathcal{L}(\mathbf{x};\boldsymbol{\theta}+\delta,p) = \mathcal{L}(\mathbf{x};\boldsymbol{\theta},\hat{p})
$:
\begin{gather}
\hat{p}(\mathbf{x}) = e^{-I(\mathbf{x},\delta)} p(\mathbf{x}),
\end{gather}
where $I(\mathbf{x},\delta):= \frac{1}{2}\mathbf{x}^\top(\delta
\mathbf{W}^\top)\mathbf{x} + \mathbf{x}^\top\delta
(\mathbf{U}^\top\mathbf{e})+C$, and $C\in\mathbb{R}$ is set to satisfy  $\displaystyle\int_{\mathbb{R}^{d}} e^{-I(\mathbf{x},\delta)}p(\mathbf{x})d\mathbf{x}=1$.}

\begin{proof}
By following \cite{GeneralizationDiffModel}, we formulate the score model $s_{\boldsymbol{\theta}}(\cdot,\cdot)$ as a random feature model:
\begin{equation}
s_{\boldsymbol{\theta}}(\mathbf{x}, t) := \frac{1}{m}\boldsymbol{\theta}\sigma(\mathbf{W}^\top\mathbf{x} + \mathbf{U}^\top\mathbf{e}_t)
\end{equation}

where $\mathbf{x}\in\mathbb{R}^{d\times 1}$,  $\boldsymbol{\theta}\in\mathbb{R}^{d\times m}$, $\mathbf{W}\in\mathbb{R}^{d\times m}$, $\mathbf{U}\in\mathbb{R}^{d_e\times m}$, $\mathbf{e}_t \in\mathbb{R}^{d_{e}\times 1}$, and $d$, $m$, $d_e$ are positive integers.

The score matching loss objective is defined as
\begin{equation} \label{Eq: score mathching loss}
    \mathcal{L}(\mathbf{x};\boldsymbol{\theta},p) := ||s_{\boldsymbol{\theta}}(\mathbf{x})-\nabla_{\mathbf{x}}\log{p(\mathbf{x})}||_{2}^{2},
\end{equation}

For the perturbation $\delta\in\mathbb{R}^{d\times m}$ in the diffusion model parameters $\boldsymbol{\theta}$, the perturbed loss value becomes:
\begin{equation}
\mathcal{L}(\mathbf{x};\boldsymbol{\theta}+\delta,p) := ||s_{\boldsymbol{\theta}+\delta}(\mathbf{x})-\nabla_{\mathbf{x}}\log{p(\mathbf{x})}||_{2}^{2}.
\end{equation}

\begin{gather}
s_{\boldsymbol{\theta}+\delta}(\mathbf{x})-\nabla_{\mathbf{x}}\log{p(\mathbf{x})} \\
= \frac{1}{m}(\boldsymbol{\theta}+\delta)\sigma(\mathbf{W}^\top\mathbf{x} + \mathbf{U}^\top\mathbf{e}) -\nabla_{\mathbf{x}}\log{p(\mathbf{x})} \\
= \frac{1}{m}\boldsymbol{\theta}\sigma(\mathbf{W}^\top\mathbf{x} + \mathbf{U}^\top\mathbf{e}) 
+ \frac{1}{m}\delta
\sigma(\mathbf{W}^\top\mathbf{x} + \mathbf{U}^\top\mathbf{e}) 
-\nabla_{\mathbf{x}}\log{p(\mathbf{x})}
\end{gather}

Let us focus on the second and third terms with the assumptions of the positive outputs for the activation function:
\begin{equation}
\frac{1}{m} \delta
(\mathbf{W}^\top\mathbf{x} + \mathbf{U}^\top\mathbf{e}) 
-\nabla_{\mathbf{x}}\log{p(\mathbf{x})}
\end{equation}

Here, let us define $I(\mathbf{x})$ as a function of $\mathbf{x}$, whose derivative is the first term of the previous equation:
\begin{equation}
\frac{\partial I(\mathbf{x})}{\partial\mathbf{x}}
= \frac{1}{m}\delta
(\mathbf{W}^\top\mathbf{x} + \mathbf{U}^\top\mathbf{e})
\end{equation}

Based on it, $I(\mathbf{x})\in\mathbb{R}$ is
\begin{equation}
I(\mathbf{x}) = \frac{1}{2}\mathbf{x}^\top(\delta
\mathbf{W}^\top)\mathbf{x} + \mathbf{x}^\top\delta
(\mathbf{U}^\top\mathbf{e})+C,
\end{equation}
with the assumption $\delta \mathbf{W}^\top$ is symmetric and where $C$ is a constant real number.

\begin{gather}
\frac{1}{m}\delta
(\mathbf{W}^\top\mathbf{x} + \mathbf{U}^\top\mathbf{e})
-\nabla_{\mathbf{x}}\log{p(\mathbf{x})} \\
= \nabla_{\mathbf{x}}I(\mathbf{x})
- \nabla_{\mathbf{x}}\log{p(\mathbf{x})} \\
= -\nabla\log{\Big(e^{-I(\mathbf{x})}p(\mathbf{x})\Big)}
\end{gather}

When $C$ is the real number that satisfies the following condition for the function $I$ with $C$:
\begin{equation}
\int_{\mathbb{R}^{d}} e^{-I (\mathbf{x})}p(\mathbf{x})=1,
\end{equation}
then we can define $\hat{p}(\mathbf{x})$ to be a perturbed PDF of inputs:
\begin{gather}
\hat{p}(\mathbf{x}) := e^{-I(\mathbf{x})} p(\mathbf{x}) \label{Eq: Perturbed distribution}\\
=\exp{\Big\{-\frac{1}{2m}\mathbf{x}^\top(\delta
\mathbf{W}^\top)\mathbf{x} - \frac{1}{m}\mathbf{x}^\top\delta
(\mathbf{U}^\top\mathbf{e})-C)\Big\}} p(\mathbf{x})
\end{gather}
\end{proof}

\textbf{Corollary 1.}\textit{
(Diffusion version of \textbf{Theorem 1})
For a given prior Gaussian distribution of noise $\epsilon\sim\mathcal{N}(0,\mathbf I)$ and the $\delta$-perturbed minimum, i.e., $\boldsymbol{\theta}+\delta$, the following $\hat{\epsilon}$ satisfies the $\mathcal{L}(\mathbf{x};\boldsymbol{\theta}+\delta,p) = \mathcal{L}(\mathbf{x};\boldsymbol{\theta},\hat{p})
$:
\begin{gather}
\hat{\epsilon} = e^{-I(\mathbf{x},\delta)} \epsilon = \mathcal{N}(\boldsymbol{\mu}_{\delta},\Sigma_\delta),
\end{gather}
where $\Sigma_{\delta}:= 
\bigg(\mathbf I + \displaystyle\frac{\delta_w}{m}\bigg)^{-1}$, $\boldsymbol \mu_{\delta}:=\displaystyle\frac{1}{m}\Sigma_{\delta}\delta_u$.
}
\begin{proof}
We provide the theoretical link that the model satisfying the $\mathcal{E}$-flat in Theorem \textbf{1} is also robust to distribution shift caused by the exposure bias.

Before that, we introduce the notations:
\begin{itemize}
    \item $\epsilon(\mathbf{x})$: the true Gaussian distribution that is known in the training process.
    \item $\hat{\epsilon} (\mathbf{x})$: the perturbed distribution that caused by the $\delta$ model perturbation in Eq. \eqref{Eq: Perturbed distribution}.
\end{itemize}

When we train the diffusion model, we add the noise $\boldsymbol \epsilon$ in the forward process and want the diffusion model to predict the $\boldsymbol \epsilon$ in the reverse process where $\boldsymbol \epsilon$ follows the normal Gaussian distribution, i.e., $\epsilon\sim \mathcal{N}(0, \mathbf I)$.
Therefore, the distribution that the model trains is the normal Gaussian, and we can define the perturbed Gaussian distribution as follows:
\begin{align}
    \hat{\epsilon} (\mathbf x)&:=e^{-I(\mathbf{x},\delta)} \epsilon \label{Eq: p*} \\
    &=\exp \bigg( -\frac{1}{2m} \mathbf x^\top \delta_w \mathbf x - \frac{1}{m}\mathbf x^\top \delta_u - C \bigg)\cdot 
    \frac{1}{\sqrt{(2\pi)^d}}\exp\bigg(-\frac{1}{2}\mathbf x^\top\mathbf x \bigg), \\
    & \qquad \text{ where } \mathbf \delta_w:= \delta \mathbf W^\top, \delta_u:= \delta \mathbf U^\top \mathbf{e}, \text{ and } \epsilon \sim \mathcal{N}(0, \mathbf I) \nonumber \\
    &= \frac{1}{\sqrt{(2\pi)^d}} \exp \bigg( -\frac{1}{2} \mathbf x^\top  (\mathbf I + \frac{1}{m}\mathbf \delta_w )\mathbf x - \frac{1}{m}\mathbf x^\top \delta_u - C \bigg)\\
    &= \frac{1}{\sqrt{(2\pi)^d}} \exp \bigg( -\frac{1}{2} \mathbf x^\top  (\mathbf I + \frac{1}{m}\mathbf \delta_w )\mathbf x - \frac{1}{m}\mathbf x^\top \delta_u - \frac{1}{2m^2}\delta_u^\top\Sigma_\delta \delta_u - \frac{1}{2}\log |\Sigma_\delta| \bigg),\\
    &\qquad \text{ where } C=\frac{1}{2m^2}\delta_u\Sigma_\delta \delta_u + \frac{1}{2}\log |\Sigma_\delta| \\
    &= \frac{1}{\sqrt{(2\pi)^d|\Sigma_\delta|}} \exp \bigg( -\frac{1}{2} 
    (\mathbf{x}- \boldsymbol\mu_\delta)^\top {\Sigma_\delta}^{-1}(\mathbf{x}- \boldsymbol\mu_\delta )
    \bigg), \text{ where } {\Sigma_\delta}:= \bigg(\mathbf I + \frac{\delta_w}{m} \bigg)^{-1}, \text{ and } \boldsymbol \mu_\delta:=-\frac{1}{m}\Sigma_\delta \delta_u 
\end{align}

Because the $\hat{\epsilon}(\mathbf x)$ is also the Gaussian distribution, we present the KL Divergence between $\hat{\epsilon}(\mathbf x)=\mathcal{N}(\boldsymbol \mu_\delta, \Sigma_\delta)$ and ${\epsilon}(\mathbf x)=\mathcal{N}( 0, \mathbf I)$ as follows:
\end{proof}

\textbf{Definition 3.}\textit{
(A set of perturbed distribution)
For a given distribution of $p(\mathbf{x})$, a set of distributions $\hat{\mathcal{P}}(\mathbf{x};p,\Delta)$ is defined as the set of perturbed distributions $\hat{p}(\mathbf{x})$:
\begin{gather}
\hat{\mathcal{P}}(\mathbf{x};p,\Delta) := \{ e^{-I(\mathbf{x},\delta)} p(\mathbf{x}) \:|\: \|\delta\|_{2}\leq\Delta \}.
\end{gather}
}

\textbf{Proposition 1.}\textit{
(A link from $\Delta$-flatness to $\hat{\mathcal{P}}$) A $\Delta$-flat minimum $\boldsymbol{\theta}^*$ achieves the flat loss values for all distributions sampled from the set of perturbed distribution:
\begin{gather}
\forall \: p\sim\hat{\mathcal{P}}(\mathbf{x};p,\Delta), \:\: \mathcal{L}(\mathbf{x};\boldsymbol{\theta}^*,p) = l^* \\
\exists \: p\nsim\hat{\mathcal{P}}(\mathbf{x};p,\Delta), \:\: \mathcal{L}(\mathbf{x};\boldsymbol{\theta}^*,p) > l^*.
\end{gather}
}

\textbf{Theorem 2.}\textit{
(A link from $\Delta$-flatness to $\mathcal{E}$-gap robustness)
A $\Delta$-flat minimum achieves $\mathcal{E}$-distribution gap robustness, such that $\mathcal{E}$ is upper-bounded as follows:
\begin{gather}
\mathcal{E}\leq \max_{\hat{p}\sim\hat{\mathcal{P}}(\mathbf{x};p,\Delta)}D(p||\hat{p}).
\end{gather}
}

\textbf{Corollary 2.}\textit{
(Diffusion version of \textbf{Theorem 2})
For a diffusion model, a $\Delta$-flat minimum achieves $\mathcal{E}$-distribution gap robustness, such that $\mathcal{E}$ is upper-bounded as follows:
\begin{align}
\mathcal{E} &\leq \max_{\|\delta\|_2\leq \Delta} \frac{1}{2} \Bigg[ \log|\Sigma_\delta | -d + tr(\Sigma_\delta^{-1}) + {\boldsymbol\mu_\delta}^\top   \Sigma_\delta^{-1} \boldsymbol\mu_\delta \Bigg]\\
&\leq \frac{1}{2}\Bigg[ \sum_i^d ( \sigma_i  - \log \sigma_i ) - d 
+ \frac{\sigma_{d}}{m^2}\|\mathbf{U}^\top\mathbf{e}\|_2^2\Delta^{2}\Bigg] 
\end{align}
where $\sigma_i$ is an eigenvalue of $\Sigma_{\delta}^{-1}$ with the increasing order of $\sigma_{1} \,\le\, \sigma_{2} \,\le\, \dots \,\le\, \sigma_{d}$.
}

\begin{proof}
From the definition \textbf{2}, a minimum $\theta^*$ hold following:
$$\forall \: \hat{p}(\mathbf{x}) \text{ s.t. } D(p||\hat{p}) \leq \mathcal{E}, \:\: \mathcal{L}(\mathbf{x};\boldsymbol{\theta}^{*},\hat{p}) = l^*. $$
Let $\sigma_i$ is an eigenvalue of $\Sigma_\delta^{-1}$ with the increasing order $\sigma_{1} \,\le\, \sigma_{2} \,\le\, \dots \,\le\, \sigma_{d}$. Then, the Diffusion version of Theorem \textbf{2} is represented as follows:
\begin{align}
\mathcal{E} &\leq \max_{\|\delta\|_2\leq \Delta} D_{KL}( \epsilon \| \hat{\epsilon} )\\
&= \max_{\|\delta\|_2\leq \Delta} \frac{1}{2} \Bigg[ \log|\Sigma_{\delta}| -d + tr(\Sigma_{\delta}^{-1}) + {\boldsymbol\mu_{\delta}}^\top  {\Sigma_{\delta}}^{-1}\boldsymbol\mu_{\delta}
\Bigg]\\
&= \max_{\|\delta\|_2\leq \Delta} \frac{1}{2}\Bigg[ \sum_i^d\log \frac{1}{\sigma_i} - d + \sum_i^d  \sigma_i  +
{\boldsymbol\mu_{\delta}}^\top  {\Sigma_{\delta}}^{-1}\boldsymbol\mu_{\delta}\Bigg] \\ 
&\leq \frac{1}{2}\Bigg[ \sum_i^d ( \sigma_i  - \log \sigma_i ) - d 
+ \frac{\sigma_{d}}{m^2}\|\mathbf{U}^\top\mathbf{e}\|_2^2\Delta^{2}\Bigg]  \label{thm2:UpperBound},
\end{align}
where inequality Eq. \eqref{thm2:UpperBound} holds when  $\boldsymbol{\mu}_\delta$ is the eigenvector satisfying $\Sigma_\delta^{-1}\boldsymbol{\mu}_\delta=\sigma_d\boldsymbol{\mu}_\delta$.
\end{proof}
\clearpage

\section{Additional experimental results}
\label{appendix:additional_experiments}
\begin{figure}
    \centering
    \includegraphics[width=0.46\linewidth]{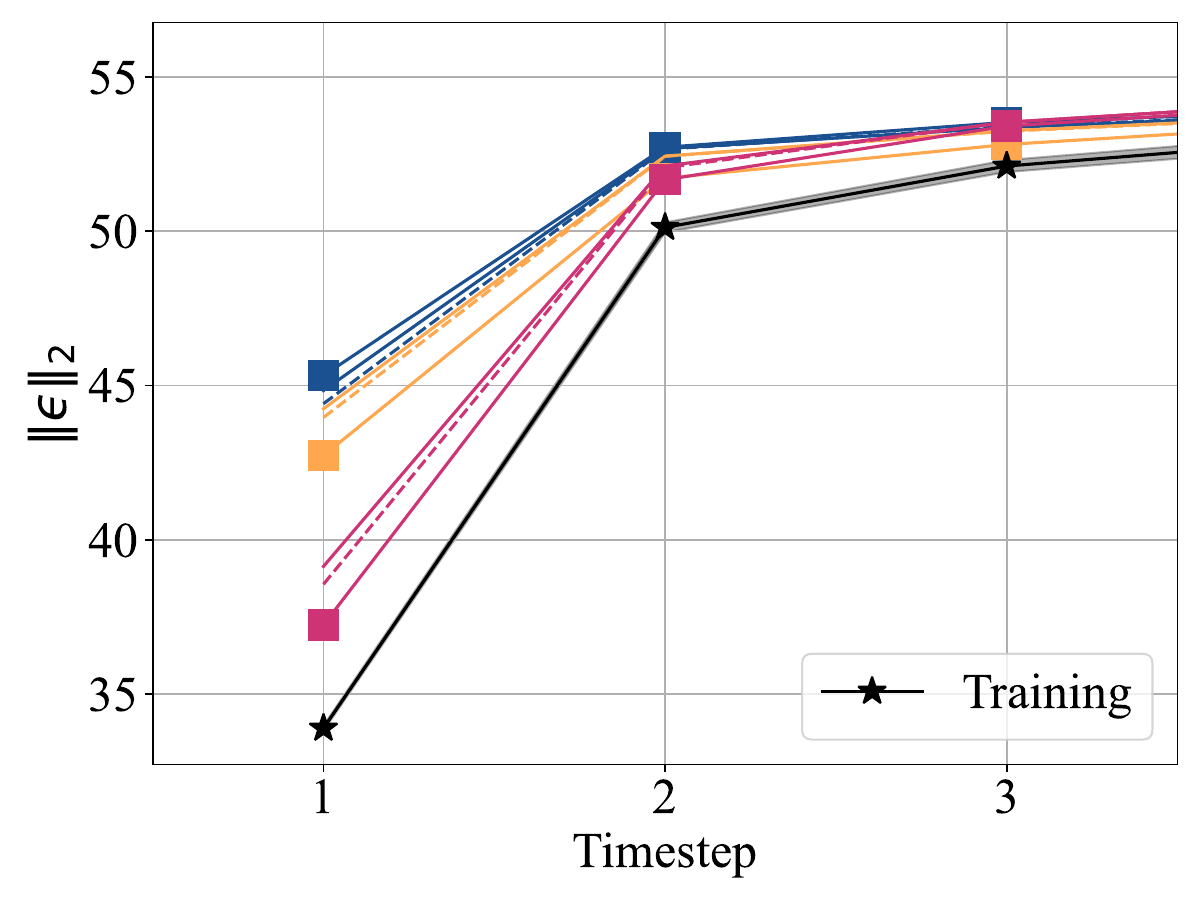}    
    \includegraphics[width=0.46\linewidth]{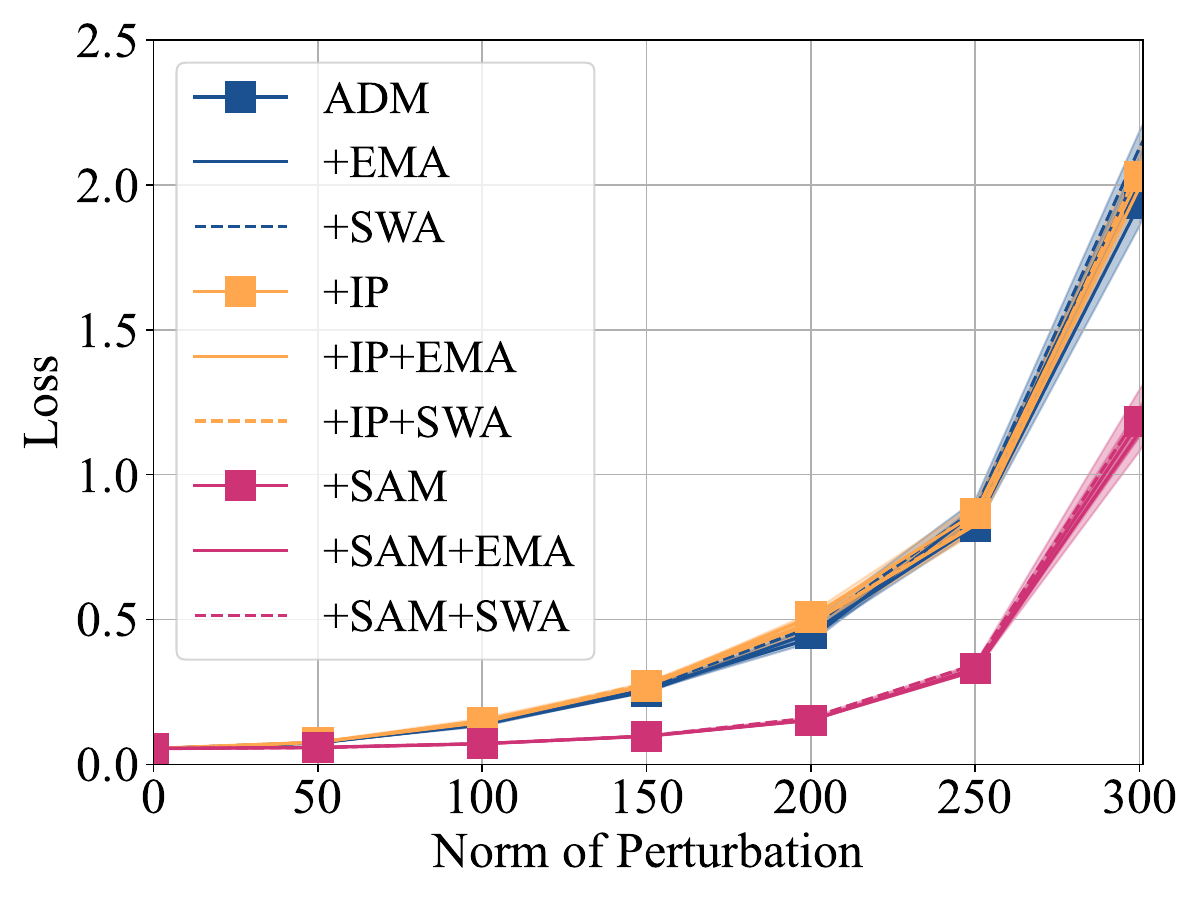}
    \caption{Additional results for CIFAR-10. We measure the L2 norm of predicted noise and loss plots under perturbation for all algorithms including \texttt{+IP+EMA}, \texttt{+IP+SWA}, \texttt{+SAM+EMA}, \texttt{+SAM+SWA}.
    }
    \label{Fig: Add-CIFAR}
\end{figure}

\begin{figure}
    \centering
    \includegraphics[width=0.48\linewidth]{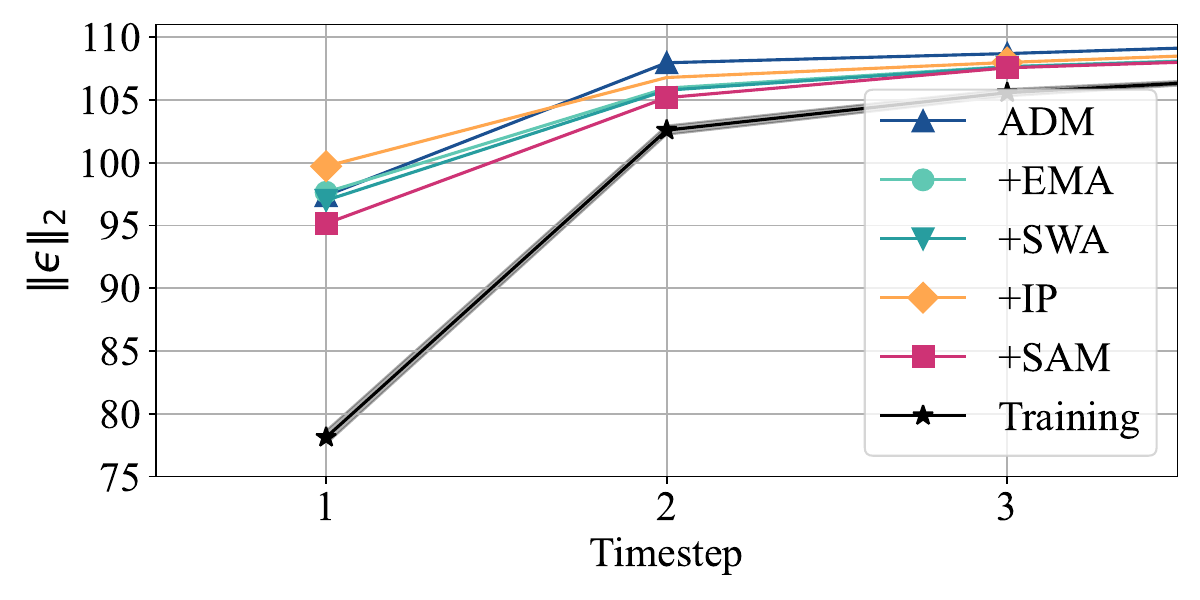}
    \includegraphics[width=0.48\linewidth]{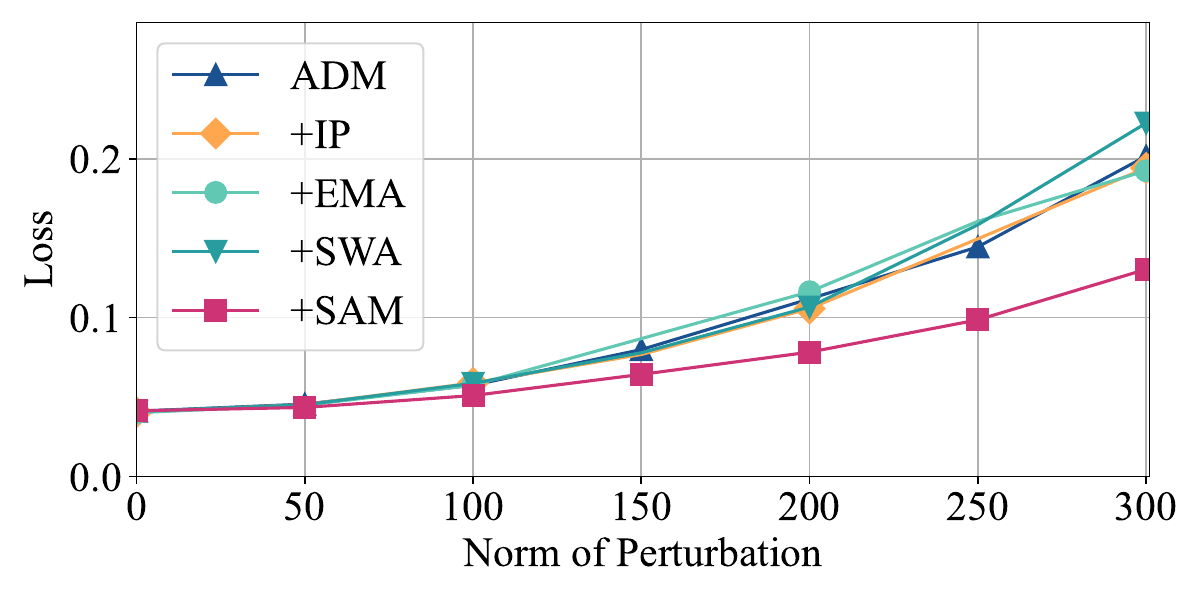}
    \caption{(Left)L2 norm of the predicted noise for LSUN Tower dataset, (Right) Loss plots under perturbation for LSUN Tower.}
    \label{Fig: LSUN-Flat}
\end{figure}

\begin{table}[t!]
\centering
\begin{tabular}{c | c c c }
\toprule
            LPF $\downarrow$ & \texttt{w/o} & \texttt{+EMA} & \texttt{+SWA}\\ \hline
            ADM & 0.091& 0.090& 0.092 \\
            \texttt{+IP} & 0.089& 0.092& 0.097 \\
            \texttt{+SAM} & \textbf{0.072} & \textbf{0.070}& \textbf{0.071} \\
\bottomrule
		\end{tabular}
    \footnotesize{
		\\ $\downarrow$: a lower value is preferred.}
\caption{Flatness measure on LSUN Tower. We calculate the loss with the perturbed model with Gaussian noise.
Lower values indicate a flatter loss landscape.
}\label{Tab: LPF-LSUN}
\end{table}
\subsection{Further results for mixture of baselines}
\label{appendix:further_results}
In Fig. \ref{Fig: Add-CIFAR}, we report additional results for \texttt{+IP+EMA, +IP+SWA, +SAM+EMA, +SAM+SWA} for CIFAR-10.
We observe that ADM already possesses a certain level of flatness supporting \texttt{+SWA} and \texttt{+EMA} fail to induce additional flatness.
We also report L2 norm of predicted noise loss plots under perturbation in Fig. \ref{Fig: LSUN-Flat} and LPF flatness in Table. \ref{Tab: LPF-LSUN} for LSUN Tower dataset.
It coincides with the result of CIFAR-10 that \texttt{+SAM} induces the lower exposure bias and flatter minima.

\subsection{Sampling results using fixed random seed}
\label{appendix:fixed_seed}
\begin{figure}[t!]
    \centering
    \includegraphics[width=0.7\linewidth]{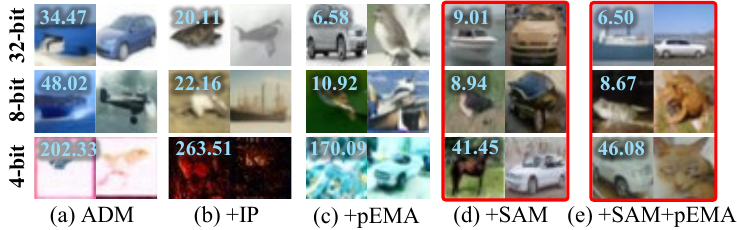}
    \caption{FID under quantization with fixed random seed.}
    \label{fig:rebuttal_fix_randomseed}
\end{figure}
We did not fix the random seed in Fig. 4 of the main paper. We report FID under varying quantization levels with a fixed seed in Fig.~\ref{fig:rebuttal_fix_randomseed}. As shown, the results remain consistent with our main findings, and all randomly selected samples follow the same trends discussed in the paper.

\subsection{Post-hoc EMA}
\label{appendix:posthoc_EMA}
\begin{table}[t!]
\centering
\resizebox{0.4\linewidth}{!}{
    \begin{tabular}{c cc}
        \toprule
        \textbf{Dataset} & \multicolumn{2}{c}{\textbf{CIFAR-10 (32x32)}}\\ 
        \hhline{|-|-|-|}
        \textbf{$T'$}
        & 20 steps 
        & 100 steps \\ \hline
        ADM+pEMA & 6.58 & 7.39 \\
        \textbf{\texttt{SAM}+pEMA} & 6.50 & 5.46 \\
        \bottomrule
    \end{tabular}
    }
    \caption{
    FID of post-hoc EMA (\texttt{+pEMA}) with ADM, \textbf{\texttt{+SAM}}.}
    \label{tab:post_ema_FID}
\end{table}
EMA provides significant performance improvements with a simple approach, but it requires cumbersome accumulation of checkpoints during training, and it is also hard to obtain a new combination of EMA after training is completed. Post-hoc EMA (pEMA) defines averaged coefficients as a power function at time $t$:
\begin{equation} \label{Eq: post-hoc EMA}
    \hat{\theta}_{\gamma}(t)=\frac{\gamma+1}{t^{\gamma+1}}\int_{0}^{t}\tau^{\gamma}\theta(\tau)d\tau,
\end{equation}
where constant $\gamma$ controls the sharpness of merged checkpoints and $\tau^{\gamma}$ determines the time weighting. $\hat{\theta}_{\gamma}(t)$ is updated as follows:
\begin{equation} \label{Eq: post-hoc EMA}
    \hat{\theta}_{\gamma}(t)=\beta_{\gamma}(t)\hat{\theta}_{\gamma}(t-1)+(1-\beta_{\gamma}(t))\theta(t), \quad \beta_{\gamma}(t)=(1-\frac{1}{t})^{\gamma+1},
\end{equation}
which is quite similar to conventional EMA.
Tab.~\ref{tab:post_ema_FID} and Fig.~\ref{fig:rebuttal_fix_randomseed} (c), (e) show that FID performance and sample visualization of post-hoc EMA(pEMA). Compared with Table 2 in the main paper, pEMA achieves the highest FID improvements; it still suffers from LPF value (0.103) even though it searches a wider range of combinations than EMA. Also, pEMA exhibits a lack of robustness, evidenced by a sharp FID degradation of $6.58\rightarrow170.09$ under 4-bit quantization. Building upon the analysis of EMA and SWA in the main paper, we argue that well-crafted weight averaging also suffers from poor flatness and robustness.

\end{document}